\documentclass[letterpaper, 10 pt, conference]{ieeeconf}  % Comment this line out if you need a4paper

\IEEEoverridecommandlockouts                              % This command is only needed if 
\usepackage[T1]{fontenc}
\usepackage{graphics} % for pdf, bitmapped graphics files
\usepackage{epsfig} % for postscript graphics files
\usepackage{mathptmx} % assumes new font selection scheme installed
\usepackage{times} % assumes new font selection scheme installed
\usepackage{amsmath} % assumes amsmath package installed
\usepackage{amssymb}  % assumes amsmath package installed

\usepackage{cite}
\usepackage{textcomp}
\usepackage{tabularx}  % 控制列宽
\usepackage{array}
\usepackage{booktabs}  % 用于更美观的表格线

\usepackage{newtxtext} 
\usepackage{multirow} % 支持合并单元格
\usepackage[font=normalsize]{caption} % 设置图注的字体为 small

\usepackage{algorithmic}
\usepackage{algorithm}

\usepackage[hidelinks]{hyperref}  % 去掉链接颜色和边框

\title{\LARGE \bf
MO-SAE:Multi-Objective Stacked Autoencoders Optimization for Edge Anomaly Detection
}

\author{Lizhao Zhang$^{1}$, Shengsong Kong$^{1}$, Tao Guo$^{1}$, Shaobo Li$^{1*}$, Zhenzhou Ji$^{1*}$% <-this % stops a space
\thanks{Corresponding authors: Shaobo Li (lishaobo@hit.edu.cn) and Zhenzhou Ji (jizhenzhou@hit.edu.cn)}% <-this % stops a space
\thanks{$^{1}$Lizhao Zhang (lizhaozhang@stu.hit.edu.cn), Shengsong Kong (shengsongkong@stu.hit.edu.cn), Shaobo Li and Zhenzhou Ji are with Faculty of computing, Harbin Institute of Technology, Harbin 150000, China.
}%
\thanks{$^{2}$Tao Guo (gt8399@hit.edu.cn) with Office of Cyber Security and Informatization, Harbin Institute of Technology, Harbin 150000, China.
        }%
}

\begin{document}

\maketitle
\thispagestyle{empty}
\pagestyle{empty}

%%%%%%%%%%%%%%%%%%%%%%%%%%%%%%%%%%%%%%%%%%%%%%%%%%%%%%%%%%%%%%%%%%%%%%%%%%%%%%%%
\begin{abstract}

Stacked AutoEncoders (SAE) have been widely adopted in edge anomaly detection scenarios. However, the resource-intensive nature of SAE can pose significant challenges for edge devices, which are typically resource-constrained and must adapt rapidly to dynamic and changing conditions. Optimizing SAE to meet the heterogeneous demands of real-world deployment scenarios, including high performance under constrained storage, low power consumption,  fast inference, and efficient model updates, remains a substantial challenge. To address this, we propose an integrated optimization framework that jointly considers these critical factors to achieve balanced and adaptive system-level optimization. Specifically, we formulate SAE optimization for edge anomaly detection as a multi-objective optimization problem and propose MO-SAE (Multi-Objective Stacked AutoEncoders).
The multiple objectives are addressed by integrating model clipping, multi-branch exit design, and a matrix approximation technique.
In addition, a multi-objective heuristic algorithm is employed to effectively balance the competing objectives in SAE optimization.
Our results demonstrate that the proposed MO-SAE delivers substantial improvements over the original approach. On the x86 architecture, it reduces storage space and power consumption by at least 50\%, improves runtime efficiency by no less than 28\%, and achieves an 11.8\% compression rate, all while maintaining application performance. 
Furthermore, MO-SAE runs efficiently on edge devices with ARM architecture. Experimental results show a 15\% improvement in inference speed, facilitating efficient deployment in cloud–edge collaborative anomaly detection systems.

\end{abstract}

%%%%%%%%%%%%%%%%%%%%%%%%%%%%%%%%%%%%%%%%%%%%%%%%%%%%%%%%%%%%%%%%%%%%%%%%%%%%%%%%
\section{INTRODUCTION}

SAE (Stacked AutoEncoders) are widely adopted in anomaly detection, particularly in edge scenarios, due to its technical merits \cite{SAE-anomaly-detect}. SAE offers efficient feature learning capabilities that enable automatic extraction of meaningful representations from raw data, and it does not rely on large amounts of labeled data. Compared to complex large-scale models, SAE operates with high efficiency, making it well-suited for long-term deployment on edge devices. However, the direct deployment of SAE presents several challenges, including operating under resource constraints, ensuring timely responses to real-time data, balancing multiple optimization objectives, and managing high communication costs, particularly in cloud–edge collaborative environments. Consequently, optimization becomes a critical step in deploying SAE for edge-based anomaly detection.

\begin{figure*}[h]
        \centering
        \includegraphics[width=\textwidth]{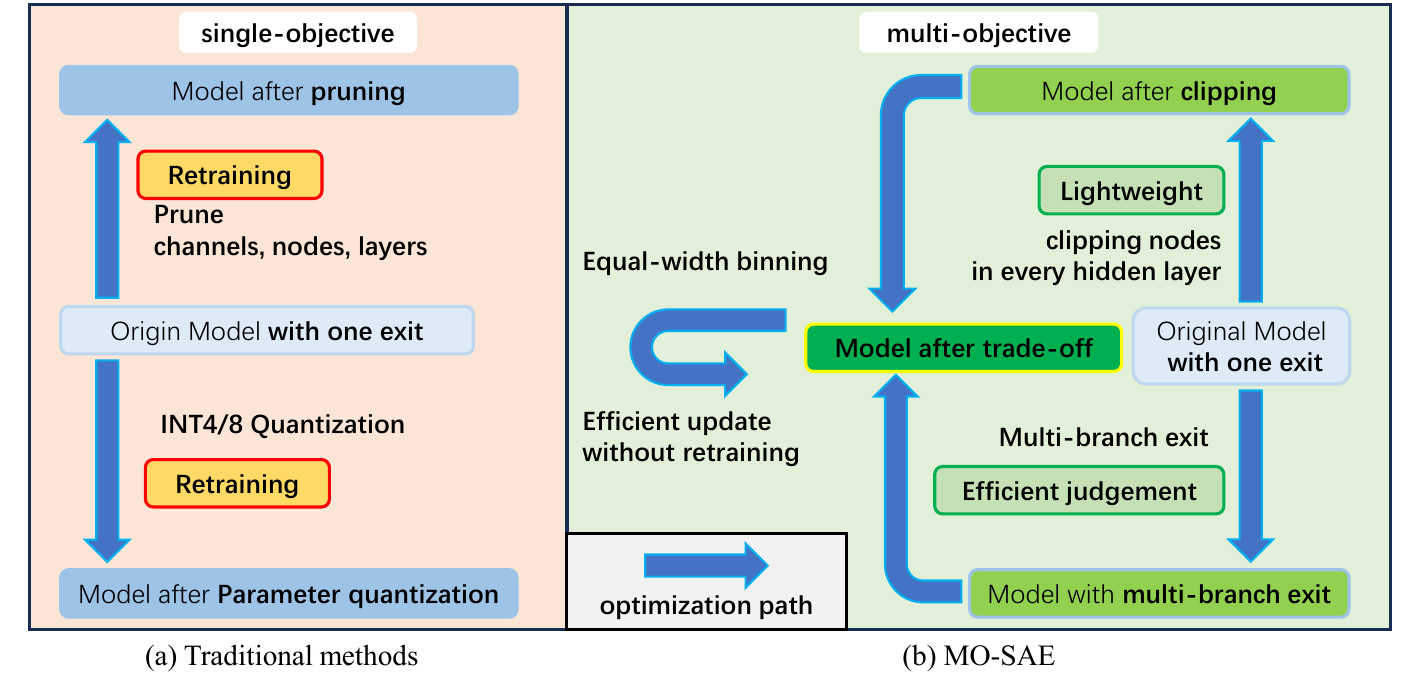} % 图片宽度设置为整页宽度
        \caption{Comparison between traditional methods and proposed MO-SAE.}
        \label{fig:1}
\end{figure*}

Researchers have conducted extensive studies on computational efficiency and optimization under resource constraints \cite{resource-constraints}, with a focus on software-based approaches \cite{BranchyNet-software3,LegoDNN-software1,Design-Space-Exploration-software2} due to the relatively high cost and complexity of hardware optimization \cite{DNN-acclerate-survey}.
\textit{Knowledge distillation} was proposed to train compact networks by transferring knowledge from a larger, fully trained network, though it poses challenges such as determining the optimal scale of the smaller network and incurring additional retraining overhead \cite{KD-anomaly-detect,knowledge-distillation-survey}. 
\textit{Weight matrix transformation} employs low-rank decomposition \cite{low-rank} to reduce storage requirements and accelerate inference, but it necessitates that the corresponding weight matrix satisfy well-defined mathematical conditions \cite{mathematical-conditions}.
\textit{Parameter sharing} \cite{Parameter-sharing} reduces model size by mapping original parameters to a constrained set using clustering \cite{Cluster} or hashing \cite{Hashing}, although it lacks comprehensive theoretical analysis \cite{theoretical-analysis}.
\textit{Parameter quantization} \cite{Parameter-quantization} replaces original parameters with low-precision counterparts to facilitate hardware acceleration \cite{Hardware-Aware-Automated-Quantization}, but its effectiveness is limited by the range of supported data types.
\textit{Model pruning}  widely used in computer vision and pattern recognition, operates at various levels \cite{vector-level,kernel-level,filter-level,Granularity-level}, yet typically requires specialized hardware and additional retraining. 
Numerous classical anomaly detection models have been introduced, including NSIBF \cite{NSIBF}, GANF \cite{GANF}, OmniAnomaly \cite{OmniAnomaly}, and LSTM \cite{LSTM}. 
However, these models typically ignore computational efficiency in resource-constrained scenarios.
Some studies \cite{Some-studies-1,Some-studies-2} propose model partitioning across edge–cloud environments to skip optimization, but this incurs significant communication overhead.

Most existing methods prioritize improving a single predefined objective, often leading to undesirable trade-offs such as increased retraining overhead or communication burden. This underscores the need for a unified approach that can effectively balance multiple competing objectives in anomaly detection. To this end, the proposed MO-SAE addresses these issues by simultaneously considering multiple optimization objectives. The core concepts include: First, based on a progressive clipping test, we propose model clipping that eliminates the retraining process. Next, a multi-branch exit design is integrated to improve runtime efficiency and generalization performance across various scenarios. Then, after formulating SAE optimization as a multi-objective problem, we introduce a heuristic algorithm to achieve trade-offs among the competing objectives. Finally, based on the theoretical analysis of approximation computation principles, we propose a matrix approximation method using equal-width binning to optimize the model update process of SAE for edge anomaly detection.

% The core concepts of the proposed framework include: First based on a progressive clipping test, this study proposes model clipping that eliminates the retraining process. Next, multi-branch exit design is integrated to improve runtime efficiency and generalization performance across various scenarios. Then, this study formulates SAE optimization as a multi-objective optimization problem and introduces a multi-objective heuristic algorithm to achieve trade-offs among competing objectives. Finally, based on the theoretical analysis of approximation computation principles, this study proposes a matrix approximation method utilizing equal-width binning to optimize the model update process of SAE for edge anomaly detection.

The key contributions of this paper are summarized as follows:

% \begin{figure*}[h]
%         \centering
%         \includegraphics[width=\textwidth]{fig1.pdf} % 图片宽度设置为整页宽度
%         \caption{Comparison between traditional methods and proposed MO-SAE.}
%         \label{fig:1}
% \end{figure*}

\begin{itemize}

\item The proposed MO-SAE aims to enhance edge anomaly detection from multiple perspectives by cohesively incorporating multiple optimization methods, including model clipping, multi-branch exit design, and matrix approximation with equal-width binning.
\item A multi-objective heuristic algorithm is employed to effectively balance multiple competing objectives in edge anomaly detection, including F1 score, runtime, storage space, and power consumption.
\item Extensive experiments are conducted on the Credit Card Fraud and SMD datasets across both x86 and ARM architectures. The results show that the proposed MO-SAE can effectively improve the F1-based accuracy, reduces latency, and lowers storage space and power consumption on edge devices.

% \item Integration of model clipping and multi-branch exit design. Based on the progressive clipping test, this study proposes model clipping, eliminating the retraining process. Furthermore, it integrates multi-branch exit design to improve runtime efficiency and generalization performance. Quantitative experiments demonstrate that SAE in edge anomaly detection exhibit strong robustness to structural perturbations and possess significant untapped optimization potential, providing a solid foundation for further exploration.
% \item A multi-objective optimization approach for SAE optimization. This study formulates SAE optimization in edge anomaly detection as a multi-objective optimization problem and introduces the multi-objective heuristic algorithm to balance multiple optimization objectives. Experimental results validate the effectiveness of our approach, showing its ability to dynamically adapt to changing real world.
% \item Model update optimization based on matrix approximation techniques. This study proposes a matrix approximation method utilizing equal-width binning to optimize the model update process by delving deeply into the approximation techniques principles underlying parameter quantization. The feasibility of the proposed method is demonstrated through both theoretical analysis and practical validation.

\end{itemize}

\section{MOTIVATION}

%In recent years, the emergence of large-scale language models, deep learning has entered an era characterized by ultra-large parameter scales and high computational costs. DeepSeek represents a significant breakthrough in optimizing ultra-large-scale Transformer models. Challenges associated with training and inference optimization continue to drive researchers to develop more efficient techniques. Similarly, in the field of anomaly detection, deep learning models face comparable challenges. Although SAE serves as a foundational approach for anomaly detection, its practical application faces several critical challenges.
Edge anomaly detection plays a crucial role in a wide range of real-world applications, such as financial fraud prevention, and real-time monitoring. SAE have emerged as a powerful unsupervised method due to their ability to learn compact representations of normal data and effectively identify anomalies. Although SAE serves as a foundational approach for anomaly detection, its practical application faces several critical challenges.
%Challenges associated with training and inference optimization drive researchers to develop more efficient techniques.

\begin{itemize}

\item Edge devices typically have constrained computational power, storage capacity, and battery life. SAEs contain a large number of parameters, demanding significant storage and computational resources.

\item Edge anomaly detection requires real-time or near-real-time anomaly detection capabilities. SAE employ deep neural network architectures to compute reconstruction errors layer-by-layer, which results in significant computational delay during inference.%, increasing the difficulty of achieving real-time anomaly detection.

\item Addressing above challenges requires effective trade-offs among multiple optimization objectives, such as reducing resource consumption without significantly sacrificing application performance.

\item Additionally, in scenarios like cloud-edge collaboration, SAE updating process is inevitable. Therefore, enhancing model update efficiency is a key challenge in SAE deployment on edge as well.

\end{itemize}

To address these challenges, this paper proposes MO-SAE, a multi-objective SAE optimization framework that minimizes resource consumption, enhances real-time detection, and improves model update efficiency for practical cloud-edge anomaly detection.

%To tackle these challenges, this paper proposes MO-SAE to formulate the SAE optimization for edge anomaly detection as a multi-objective optimization problem and integrates model clipping, multi-branch exit design, the heuristic algorithm and matrix approximation technique to improve computational efficiency, minimize resource consumption, enhance real-time detection capabilities and model update efficiency for practical application of anomaly detection SAE especially in cloud-edge collaborative scenarios.
% To tackle these challenges, this paper proposes MO-SAE, a multi-objective SAE optimization framework designed to minimize resource consumption, improve computational efficiency, and enhance both real-time detection capabilities and model update efficiency, especially for practical deployment of anomaly detection SAE in cloud-edge collaborative scenarios. MO-SAE adopt model clipping to reduce storage space without retraining, integrate multi-branch exit design to improve runtime efficiency and real-time detection capabilities with better generalization, introduce the heuristic algorithm to address the multi-objective SAE optimization problem, and optimize model update process by matrix approximation technique.

\section{METHOD AND FRAMEWORK}

In this section, we present a detailed description of MO-SAE in the following three parts. First, MO-SAE is proposed by integrating model clipping, multi-branch exit design, and a matrix approximation technique, followed by in-depth comparisons with traditional methods. Second, we formulate the edge SAE as a multi-objective optimization problem and introduce a heuristic algorithm to balance competing objectives. Third, we utilize a matrix approximation technique to optimize the model update process. This is achieved by applying equal-width binning and approximate reconstruction to the original parameter matrix.

% In this section, we elaborate on MO-SAE through the following parts. First, by conducting in-depth comparisons with traditional methods, this study proposes MO-SAE based on model clipping, multi-branch exit design, and matrix approximation technique. Second, the edge SAE is formulated as a multi-objective optimization problem, introducing the heuristic algorithm to achieve trade-offs among multiple optimization objectives. Third, this study utilizes matrix approximation technique to optimize the model update process by applying equal-width binning and approximate reconstruction to the original parameter matrix values.

\subsection{Outline of Optimization Framework} \label{Outline of Integrated Optimization Approach} 
Whether retraining is necessary for pruned deep learning models remains a debated issue in model optimization. For instance, models that are highly robust to structural perturbations may require less or no additional retraining. However, many existing studies in model optimization do not evaluate a model's robustness prior to optimization.
% During model optimization, the necessity of retraining for all deep learning models after pruning remains a topic of debate. For instance, models with strong robustness to structural perturbations can effectively reduce extra retraining overhead. However, many current studies on model optimization fail to assess in advance whether the model to be optimized possesses strong robustness.

% \begin{figure*}[h]
%         \centering
%         \includegraphics[width=\textwidth]{fig1.pdf} % 图片宽度设置为整页宽度
%         \caption{Comparison between traditional methods and proposed MO-SAE.}
%         \label{fig:1}
% \end{figure*}

\begin{figure}[htbp]
        % \noindent
        \centering
        \includegraphics[width=0.45\textwidth]{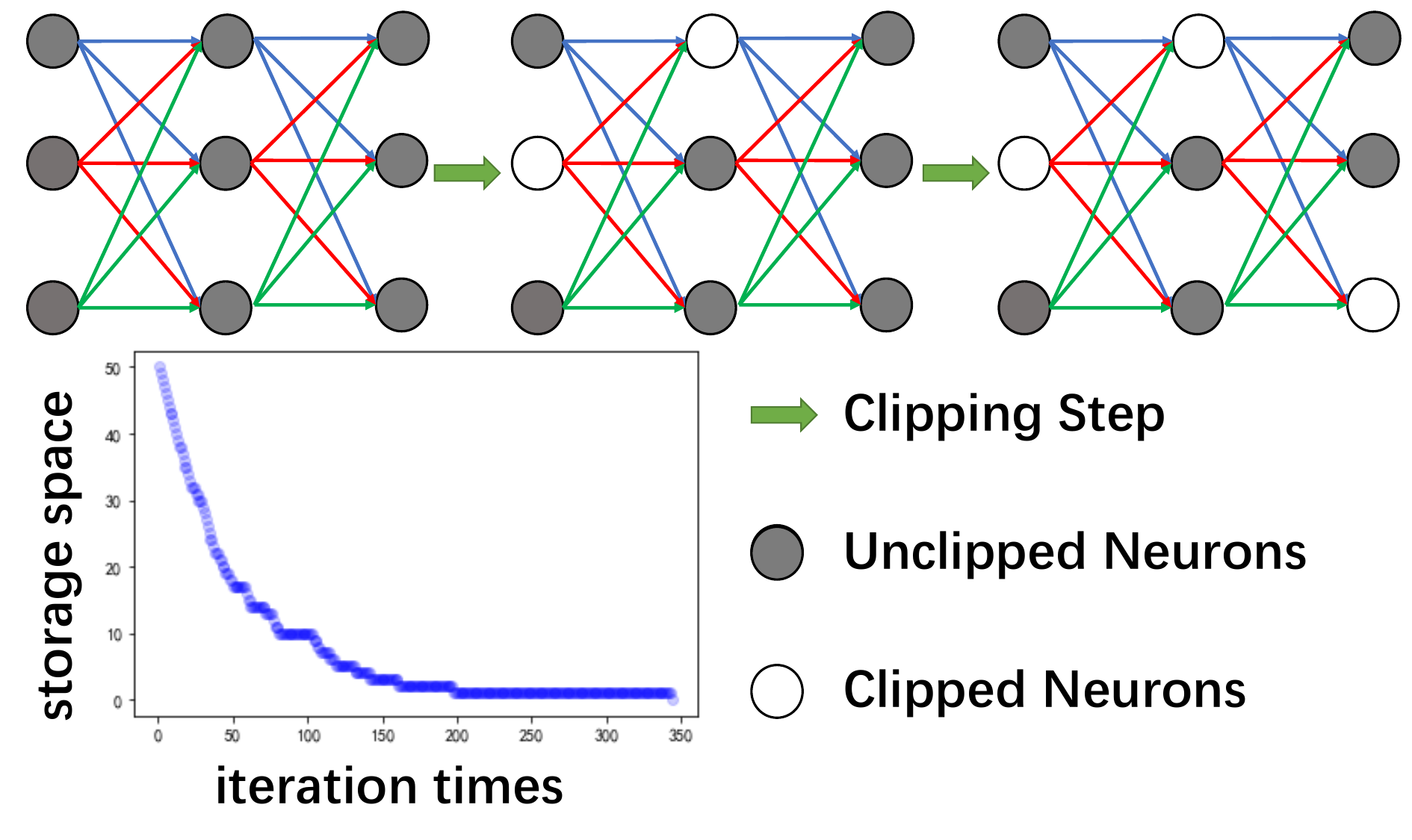}
        \caption{Progressive clipping test.}
        \label{fig:2}
        % \noindent
        \centering
        \includegraphics[width=0.45\textwidth]{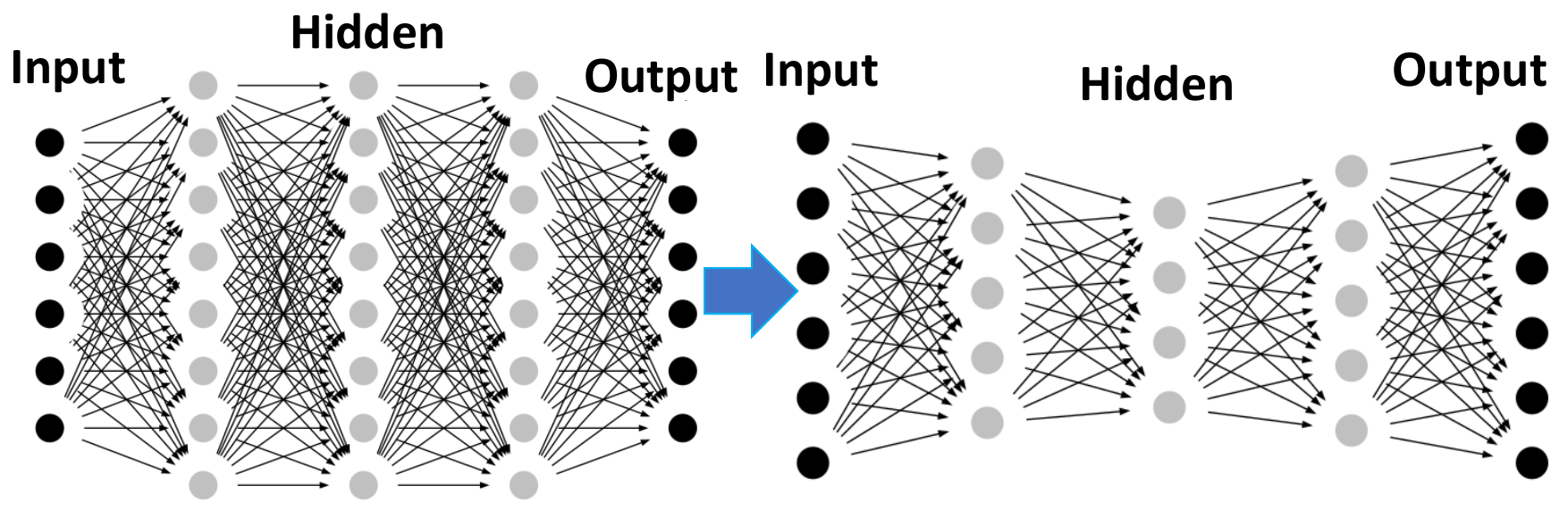}
        \caption{Model clipping design.}
        \label{fig:3}
        % \noindent
        \centering
        \includegraphics[width=0.45\textwidth]{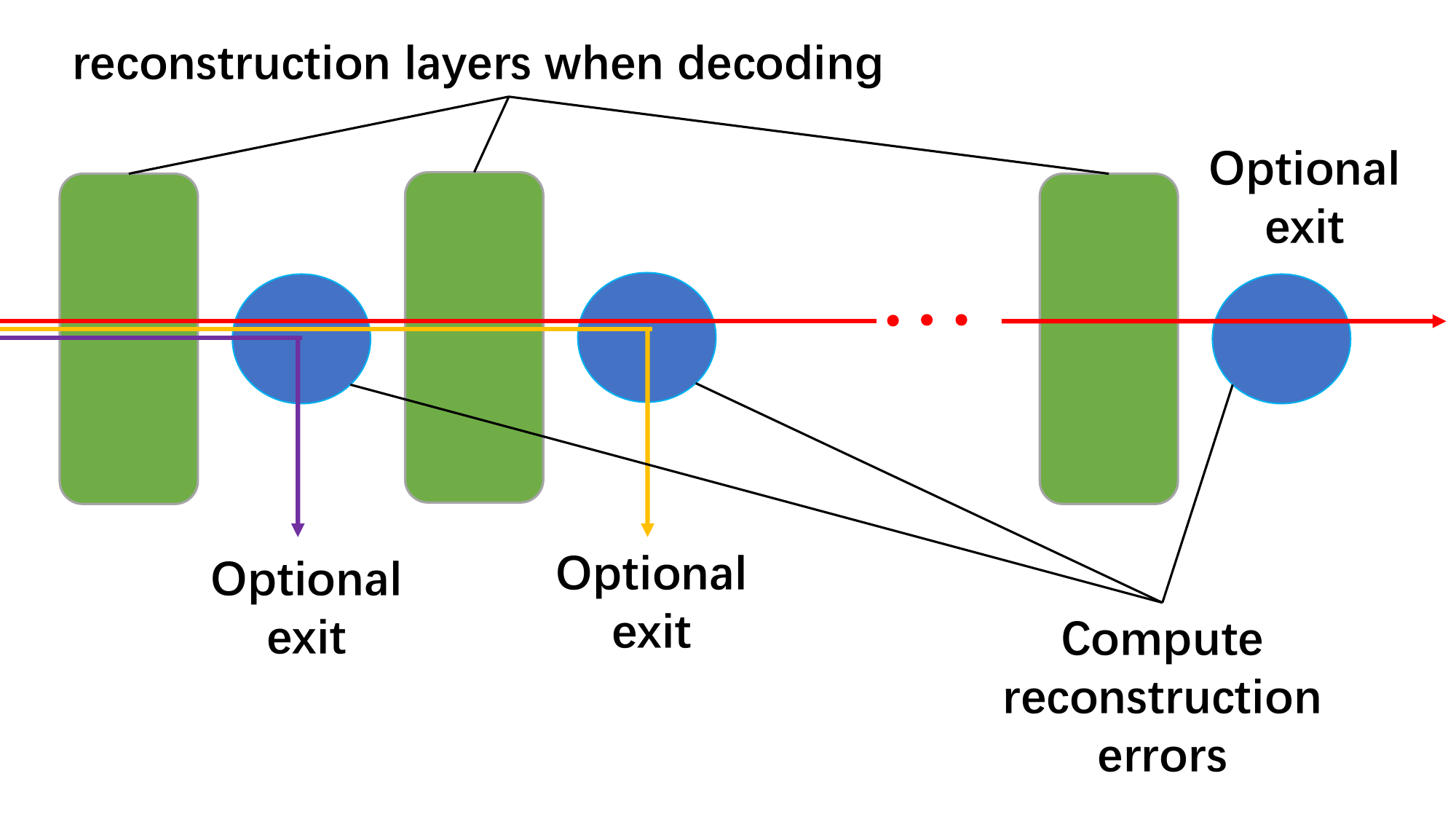}
        \caption{Multi-branch exit design.}
        \label{fig:4}
        % \noindent
        \centering
        \includegraphics[width=0.45\textwidth]{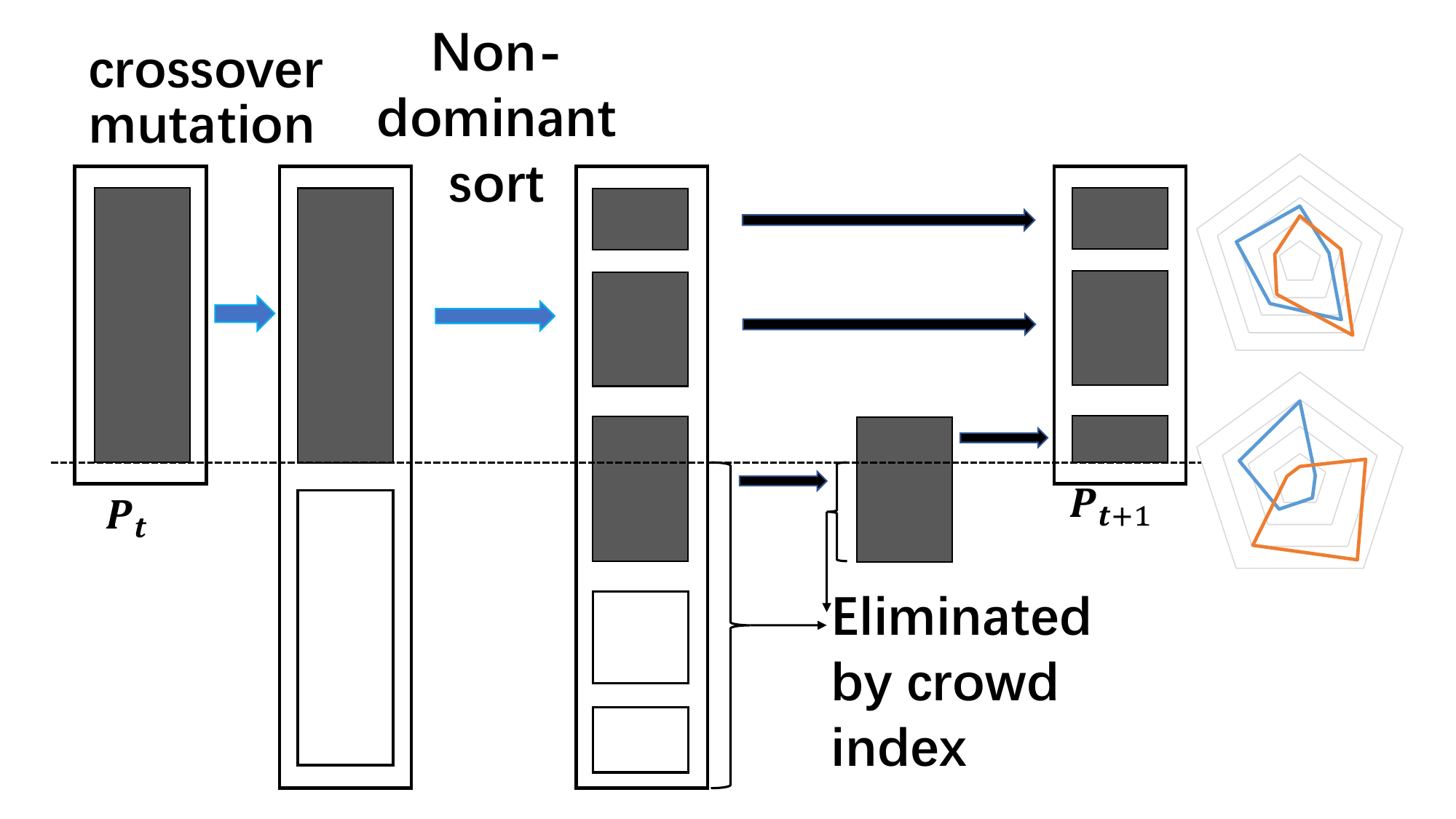}
        \caption{Multi-objective heuristic algorithm workflow.}
        \label{fig:5}
        % % \noindent
        \centering
        \includegraphics[width=0.45\textwidth]{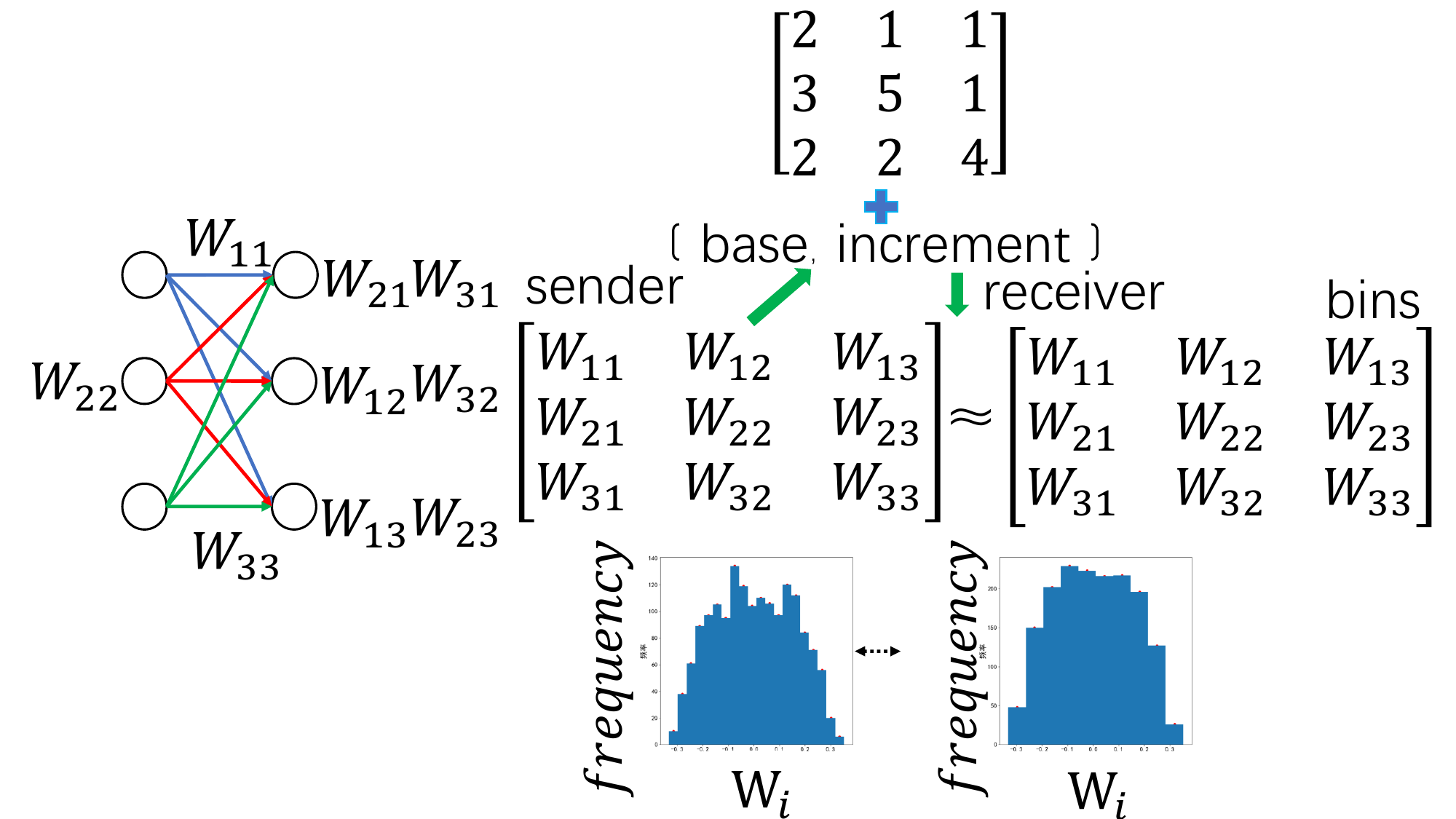}
        \caption{Model update optimization in transmission process.}
        \label{fig:6}
\end{figure}

To address this issue, we adopt a progressive clipping test to evaluate whether the SAE for edge anomaly detection exhibits strong robustness to structural perturbations. In this study, the progressive clipping test utilizes a random sampling method to select neurons. The results of successive samplings are combined to generate a series of clipping schemes, covering mild to extreme levels of pruning, as illustrated in Fig. ~\ref{fig:2}. Based on the analysis in Section ~\ref{Analysis of Optimization Potential}, we propose three techniques to improve SAE for edge anomaly detection: model clipping (which eliminates retraining), multi-branch exit design (which improves runtime efficiency and generalization performance), and matrix approximation with equal-width binning (which optimizes model update process). The effectiveness of these techniques is demonstrated through comparisons between traditional methods and MO-SAE, as shown in Fig. ~\ref{fig:1}. Specifically, the model clipping, illustrated in Fig. ~\ref{fig:3}, focuses on pruning nodes within each hidden layer rather than pruning channels or entire layers. The multi-branch exit design, illustrated in Fig. ~\ref{fig:4}, introduces multiple exits before the final output of the SAE. After data samples are fed into the model, the reconstruction error is computed at each possible exit and compared against a predefined reconstruction error threshold, thereby enhancing the runtime efficiency of edge anomaly detection. The model update optimization based on equal-width binning is further analyzed in Section ~\ref{Model Update Optimization}.

% To address this issue, we adopt the progressive clipping test to evaluate whether SAE for edge anomaly detection exhibit strong robustness to structural perturbations. In this study, the progressive clipping test utilizes a random sampling method to select neurons, with the results of successive samplings combined to generate a series of clipping schemes that can cover mild to extreme levels of clipping illustrated in Fig.~\ref{fig:2}. Based on the analysis in ~\ref{Analysis of Optimization Potential}, this study proposes model clipping that eliminates the retraining process, multi-branch exit design and model update optimization to improve SAE in edge anomaly detection from multiple aspects, as demonstrated in the comparison between traditional methods and MO-SAE illustrated in Fig.~\ref{fig:1}. Thereinto, the model clipping, illustrated in Fig.~\ref{fig:3}, focuses on pruning nodes within each hidden layer rather than pruning channels or entire layers. The multi-branch exit design, illustrated in Fig.~\ref{fig:4}, introduces multiple exits before the final output of the SAE. After data samples are fed into the model, the reconstruction error is computed at each possible exit and compared against a predefined reconstruction error threshold, thereby enhancing the runtime efficiency of edge anomaly detection. The model update optimization based on equal-width binning will be thoroughly analyzed and discussed in D.

% \begin{figure}[htbp]
%         % \noindent
%         \centering
%         \includegraphics[width=0.45\textwidth]{fig3.pdf}
%         \caption{Model clipping design.}
%         \label{fig:3}
% \end{figure}
% \begin{figure}[htbp]
%         \centering
%         \includegraphics[width=0.45\textwidth]{fig4.pdf}
%         \caption{Multi-branch exit design.}
%         \label{fig:4}
% \end{figure}

\subsection{Formulation and Algorithm} 
Our objective is to ensure that the optimized SAE delivers high application performance and runtime efficiency, while minimizing storage space and power consumption. Therefore, we formulate SAE optimization in edge anomaly detection as a multi-objective optimization problem and introduce a multi-objective heuristic algorithm to jointly optimize the model clipping and the multi-branch exit design, as illustrated in Algorithm \ref{alg:1} to manage trade-offs among conflicting optimization goals. The workflow is illustrated in Fig. ~\ref{fig:5}. 
The multi-objective heuristic algorithm does not aggregate multiple objectives into a single weighted function. Instead, it uses non-dominated sorting to divide the population into different levels, thereby preserving a set of solutions that reflect various trade-offs among objectives.
In brief, the multi-objective heuristic algorithm adopts the same selection, crossover, and mutation processes as the genetic algorithm, with two essential modifications. On one hand, the algorithm employs non-dominated sorting to rank individuals, ultimately generating populations with specialized capabilities to address varying task requirements. On the other hand, the selection probability is designed to satisfy two principles simultaneously. First, individuals from higher-ranked populations are more likely to be selected. Second, all individuals within the same population have an equal probability of selection. Specifically, assume that there are $k$ selected non-dominated populations, denoted as $Z_1$, $Z_2$, …, $Z_k$. Here $Z_i$ represents the $i$-th non-dominated population, and $count(Z_i)$ denotes the number of individuals in $Z_i$. Thus, the selection probability for an individual is calculated as shown in Equation \ref{eq:9}.

% Our objective is to ensure the optimized SAE deliver high application performance and runtime efficiency, while minimizing storage space and power consumption as much as possible. Therefore, this study formulates the SAE optimization in edge anomaly detection as a multi-objective optimization problem and introduces the multi-objective heuristic algorithm for model clipping and multi-branch exit design to manage competitive relationships among multiple optimization goals as illustrated in Algorithm \ref{alg:1}. The workflow is illustrated in Fig.~\ref{fig:5}. In brief, the multi-objective heuristic algorithm utilizes same selection, crossover, and mutation process as genetic algorithm with two essential modifications. On one hand, the multi-objective heuristic algorithm integrates non-dominant relationship to sort, ultimately producing populations with specialized strengths to handle varying task requirements. On the other hand, the probability of being chosen needs to guarantee 2 principles at the same time, firstly an individual assigned to a higher-level population is more likely to be selected; secondly, all individuals within the same population have an equal chance of being selected. Specifically, we can make assumptions that there are $k$ chosen non-dominant populations numbered from 1 to $i$, $Z_i$ stands for the corresponding non-dominant population, $count(Z_i) $ equals to the number of individuals in $Z_i$. Thus, the individual probability of being chosen is computed as equation \ref{eq:9}.

\vspace{-1em}
\begin{algorithm}[!h]
\small
\caption{proposed non-dominant sort}
\label{alg:1}
\renewcommand{\algorithmicrequire}{\textbf{Input:}}
\renewcommand{\algorithmicensure}{\textbf{Output:}}

\begin{algorithmic}[1]
        \REQUIRE $P$(the population set)  %%input
        \ENSURE $F_{i}$(non-dominant sets at different grade)    %%output
        
        \STATE function qdenomin(P)
        
        \FOR{every p in P}
        \STATE $S_{p} = \varnothing$
        \STATE $n_{p}$ = 0

                \FOR{every q in P}
                        \IF{q domin p}
                        \STATE then add q to $S_{p}$
                        \ELSE 
                                \IF{p domin q}
                                \STATE then $n_p=n_{p} + 1$
                                \ENDIF
                        \ENDIF
                        \IF{$n_p$ = 0}
                        \STATE $P_rank$ = 1
                        \STATE add p to $F_1$
                        \ENDIF
                \ENDFOR

        \ENDFOR
        
        \STATE i=1
        \WHILE{$F_i$ not empty}
                \STATE $Q = \varnothing$
                \FOR{every p in $F_i$}
                        \FOR{every q in $S_p$}
                                \STATE $n_p=n_{p} - 1$
                                \IF{$n_p = 0$}
                                \STATE $q_{rank} = i + 1$
                                \STATE add q to Q
                                \ENDIF
                        \ENDFOR
                \ENDFOR
                \STATE i=i+1
                \STATE $F_i=Q$
        \ENDWHILE
        \STATE end function

\end{algorithmic}
\end{algorithm}

% \begin{algorithm}[!h]
% \small
% \caption{computation of crowd index}
% \label{alg:2}
% \renewcommand{\algorithmicrequire}{\textbf{Input:}}
% \renewcommand{\algorithmicensure}{\textbf{Output:}}

% \begin{algorithmic}[1]
%         \REQUIRE $S$(the non-dominant set to be divided by crowd \\
%                 \hspace{1.2em} index; num：The number of elements in S) \\
%                 \hspace{1.2em} $T$(aspects of goals)  %%input
%         \ENSURE $S[i]_{distence}$(crowd index for everyone in S)    %%output
        
%         \STATE  function distance(S)

%         \FOR{every i in num}
%                 \STATE $S[i]_{distence} = 0$
%         \ENDFOR
        
%         \FOR{every t in T}
%                 \STATE S=sort(S,T)
%                 \STATE $S[1]_{distence} = S[count(S)]_{distence} = \infty $
%                         \FOR{every i from 2 to num-1}   
%                                 \STATE $S[i]_{\text{distance}} = S[i]_{\text{distance}} +$
%                                 \STATE $\frac{(S[i + 1].T - S[i - 1].T)}{(\max(S[i].T) - \min(S[i].T))}$
%                         \ENDFOR
%         \ENDFOR

%         \STATE end function

% \end{algorithmic}
% \end{algorithm}

% \begin{figure}[hb]
%         % \noindent
%         \centering
%         \includegraphics[width=0.45\textwidth]{fig5.pdf}
%         \caption{Multi-objective heuristic algorithm workflow.}
%         \label{fig:5}
% \end{figure}

\vspace{-1em}
\begin{equation}
\small
% \frac{\frac{1}{i}}{\sum_{i=1}^k \frac{1}{i} \cdot \text{count}(Z_i)} \label{eq:9}
{\frac{1}{i}} / ({\sum_{i=1}^k \frac{1}{i} \cdot \text{count}(Z_i)}) \label{eq:9}
\end{equation}
\vspace{-1em}

We develop separate genetic information modules for the model clipping and the multi-branch exit design. For model clipping, the genetic information is encoded at the granularity of individual computational nodes in the hidden layers. Nodes marked for pruning are represented by 0, while those retained are represented by 1. For multi-branch exit design, the IEEE 754 floating-point format is adopted to encode different threshold percentages for reconstruction errors at each exit. For the combined case of the model clipping and the multi-branch exit design, the genetic information is formed by concatenating the two modules described above.

% This study developed the genetic information module for model clipping and multi-branch exit design, respectively. For model clipping, the genetic information module is encoded at the granularity of individual computational nodes in the hidden layers, cut marked as 0, else as 1. For multi-branch exit design, generalizable IEEE754 is adopted to encode different exit percentage thresholds of reconstruction errors. And for model clipping combined with multi-branch exit design, the genetic information module is concatenated from the above 2 genetic information modules.

\subsection{Theoretical Analysis of Model Update Optimization} \label{Model Update Optimization}

% Approximate computing is a strategy that trades precision for efficiency within an acceptable error range, striving to balance accuracy, efficiency, and resource utilization. This approach is especially beneficial for ultra-low-power devices or scenarios with constrained storage. We employ matrix approximation to optimize the model update process. Specifically, equal-width binning is applied to the original parameter matrix, followed by approximate reconstruction.

% Approximate computing is a strategy that trades precision for efficiency within an acceptable error range, striving to balance accuracy, efficiency, and resource utilization. This approach is especially beneficial for ultra-low-power devices or scenarios with constrained storage. \iffalse Model quantization, as an implementation of approximate computing, is often constrained by hardware designs, typically supporting only int4, int8, or int16 formats. These fixed formats are not flexible enough to accommodate the distribution of model parameters, requiring retraining to minimize quantization errors, struggling to achieve finer trade-offs among storage, energy consumption, and accuracy—a requirement for SAE deployment in edge anomaly detection.\fi This study employs matrix approximation to optimize model updates process by applying equal-width binning to the original parameter matrix, followed by approximate reconstruction.

Approximate computing is a strategy that trades precision for efficiency within an acceptable error range, striving to balance accuracy, efficiency, and resource utilization. Specifically, we employ equal-width binning from matrix approximation to optimize the model update process. 
As illustrated in Fig. ~\ref{fig:6}, the model sender first determines the bin density—the number of partitions. The width of each bin is calculated based on the minimum and maximum values of the matrix. Each value is then mapped to its corresponding bin based on its magnitude, and the bin is used to represent the original value. This operation transforms continuous numerical data into several equal-width discrete bin values. Finally, the sender transmits three types of parameters derived from the binning process: base, increment, and step size. On the model receiver side, these parameters are used to reconstruct an approximate version of the original parameter matrix, enabling the inference process. To evaluate the effectiveness and feasibility of the approximate matrix, this study conducted a theoretical analysis, as outlined in Equations \ref{eq:10}--\ref{eq:15}. Here, $A$ represents the original matrix, approximated as $(A + \Delta A)$, and $x$ is the true and unbiased input data, approximated as $(x + \Delta x)$. The error term $\Delta A$ is introduced by the approximate matrix, while $\Delta x$ arises from system measurement noise. In this context, $\| \|$ denotes the 2-norm, so $\| A \| \| A^{-1} \|$ is a fixed and well-defined value.

% As illustrated in Fig.~\ref{fig:6}, the model sender determines bin density (the number of partitions) at first. The width of each bin is calculated based on the minimum and maximum of the matrix. Then each value is subsequently mapped to its corresponding bin based on its magnitude, after which the bin is used to represent the original value. This operation transforms continuous numerical data into several equal-width discrete bin values. Finally, the sender transmits three types of parameters derived from the binning process, namely: base, increment, and step numbers. On the model receiver side, these parameters are used to reconstruct an approximate version of the original parameter matrix, enabling the inference process. To evaluate the effectiveness and feasibility of the approximate matrix, this study conducted theoretical analysis outlined in equation \ref{eq:10}--\ref{eq:15}. Here, $A$ represents the original matrix, approximated as $(A + \Delta A)$, and $x$ is the true and unbiased input data, approximated as $(x + \Delta x)$. $\Delta A$ is introduced by the approximate matrix, while $\Delta x$ arises from system measurement error. In this context, $\| \|$ denotes the 2-norm, so $\| A \| \| A^{-1} \|$ is a fixed and well-defined value.

% \begin{figure}[htb]
%         % \noindent
%         \centering
%         \includegraphics[width=0.45\textwidth]{fig6.pdf}
%         \caption{Model update optimization in transmission process.}
%         \label{fig:6}
% \end{figure}

A good approximation requires satisfying the condition $A x = b$, which is equivalent to $(A + \Delta A)(x + \Delta x) = b$. However, due to the inevitable presence of system measurement error $\Delta x$, the ideal case of $A x = b$ becomes unattainable. Instead, both $A(x + \Delta x) = b$ and $(A + \Delta A)(x + \Delta x) = b$ are approximations of the ideal scenario, differentiated by the distinct error terms. Based on the equivalence of $A x = b$ and $(A + \Delta A)(x + \Delta x) = b$, it can be inferred that both the approximate weight error $\Delta A$, introduced by the bin density, and the system measurement error $\Delta x$, can be treated as approximately zero matrices, thereby ensuring that the equation holds approximately. Since $\Delta x$ is determined by the state of the input data, focusing on bin density becomes a reasonable strategy for optimizing model updates. As demonstrated in the subsequent derivations of inequalities, the infinitesimal relationship between $\Delta x$ and $\Delta A$ can be established. Assuming that the high-order terms of the system measurement error $\Delta x$ are bounded, it follows that a good approximation can be achieved with a reasonable configuration of bin density for $\Delta A$.

% A good approximation requires satisfying the condition $A x = b$, which is equivalent to $(A + \Delta A)(x + \Delta x) = b$. However, due to the inevitable presence of system measurement error $\Delta x$, the ideal case of $A x = b$ is unattainable. Instead, both $A(x + \Delta x) = b$ and $(A + \Delta A)(x + \Delta x) = b$ are approximations of the ideal scenario, differentiated by the distinct error terms. Based on the equivalence of $A x = b$ and $(A + \Delta A)(x + \Delta x) = b$, reverse reasoning reveals that the approximate weight error $\Delta A$, caused by the bin density, and the system measurement error $\Delta x$ are both approximately zero matrices, ensuring that the equation holds approximately. Since $\Delta x$ is determined by the state of the input data, focusing on bin density is reasonable for model update optimization. As demonstrated in the subsequent inequality derivations, by examining the infinitesimal relationship between $\Delta x$ and $\Delta A$, and assuming that the high-order terms of the system measurement error $\Delta x$ are bounded, it follows that a good approximation can be achieved with a reasonable configuration of bin density for $\Delta A$.

\vspace{-0.7em}  % 调整为合适的值
\begin{align}
(A + \Delta A)(x + \Delta x) = b \label{eq:10}
\end{align}
\begin{align}
A \Delta x + \Delta A x + \Delta A \Delta x = 0 \label{eq:11}
\end{align}
\begin{align}
\Delta x = -A^{-1}(\Delta A)(x + \Delta x) \label{eq:12}
\end{align}
\begin{align}
\| \Delta x \| \leq \| A^{-1} \| \| \Delta A \| \| x + \Delta x \| \label{eq:13}
\end{align}
\begin{align}
% \frac{\| \Delta x \|}{\| x + \Delta x \|} \leq \frac{\| A^{-1} \| \| \Delta A \| \| A \|}{\| A \|} \label{eq:14}
\| \Delta x \| / \| x + \Delta x \| \leq \| A^{-1} \| \, \| \Delta A \| \, \| A \| / \| A \| \label{eq:14}
\end{align}
\begin{align}
% \frac{\| \Delta x \| \| A \|}{\| x + \Delta x \| \| \Delta A \|} \leq \| A \| \| A^{-1} \| \label{eq:15}
\| \Delta x \| \| A \| / {\| x + \Delta x \| \| \Delta A \|} \leq \| A \| \| A^{-1} \| \label{eq:15}
\end{align}
\vspace{+0.4em}

\section{QUANTITATIVE EXPERIMENTS}

\subsection{Datasets, Testbed and evaluation framework}

To demonstrate the effectiveness, objectivity, and generality of the integrated optimization framework, quantitative experiments were conducted on real-world anomaly detection datasets. The SMD dataset contains per-minute operational data collected from 28 servers over a period of five weeks. It comprises 1,416,825 data entries with 38 features, among which 4.16\% are labeled as anomalies. The Credit Card Fraud \cite{Credit-Card-Fraud-dataset1} dataset includes 284,807 European credit card transactions from September 2013, with 30 features and an anomaly rate of 0.17\%.

This study uses an x86-based architecture to train and evaluate the optimized SAE model. The specific configuration consists of an Intel® Core™ i7-6700HQ CPU (2.60GHz), 8 GB of DDR4 2133MHz RAM, and an NVIDIA GeForce GTX 960M GPU (2GB). The NVIDIA\textsuperscript{\textregistered} Jetson Xavier\textsuperscript{\texttrademark}, built on the ARM architecture, offers configurable modes with multiple power and core options. It is commonly used as an edge device. Thus, employing the NVIDIA\textsuperscript{\textregistered} Jetson Xavier\textsuperscript{\texttrademark} to validate the effectiveness of the solutions derived from the proposed optimization framework is well-suited for edge deployment scenarios.

To ensure a comprehensive evaluation of SAE in edge anomaly detection, the evaluation framework should be both generalizable and capable of demonstrating the effectiveness of optimization methods across diverse architectural platforms. This is achieved by using relative metrics rather than relying on absolute values. More unclipped neurons imply greater storage requirements; more basic operations lead to higher power consumption; and shorter runtime indicates better runtime efficiency. To evaluate the impact of model clipping and the multi-branch exit design, this study adopts three metrics—namely: the classic F1 score, running time, and compression rate—to assess application performance, runtime efficiency, and model update efficiency, respectively. Specifically, the compression rate is defined in Equation \ref{eq:16}, where $N$ represents the total number of parameter values. Maximum storage space and power consumption are normalized to enable comparison using relative values.

\vspace{-0.3em}
\begin{equation}
        % \text{compression rate} = \frac{\left(\left(\frac{\log_2 d_x}{8}\right) \cdot N + 2 \cdot 8\right)}{8 \cdot N} \label{eq:16}
\text{compression rate} = \left(\left(\log_2 d_x / {8}\right) \cdot N + 2 \cdot 8\right) / {8 \cdot N} \label{eq:16}
\end{equation}
\vspace{-1em}

% \begin{align}
% \text{compression rate} = \frac{\left(\left(\frac{\log_2 d_x}{8}\right) \cdot N + 2 \cdot 8\right)}{8 \cdot N} \label{eq:16}
% \end{align}

\subsection{Analysis of Optimization Potential} \label{Analysis of Optimization Potential}

\begin{figure}[ht]
        \centering
        \includegraphics[width=0.46\textwidth]{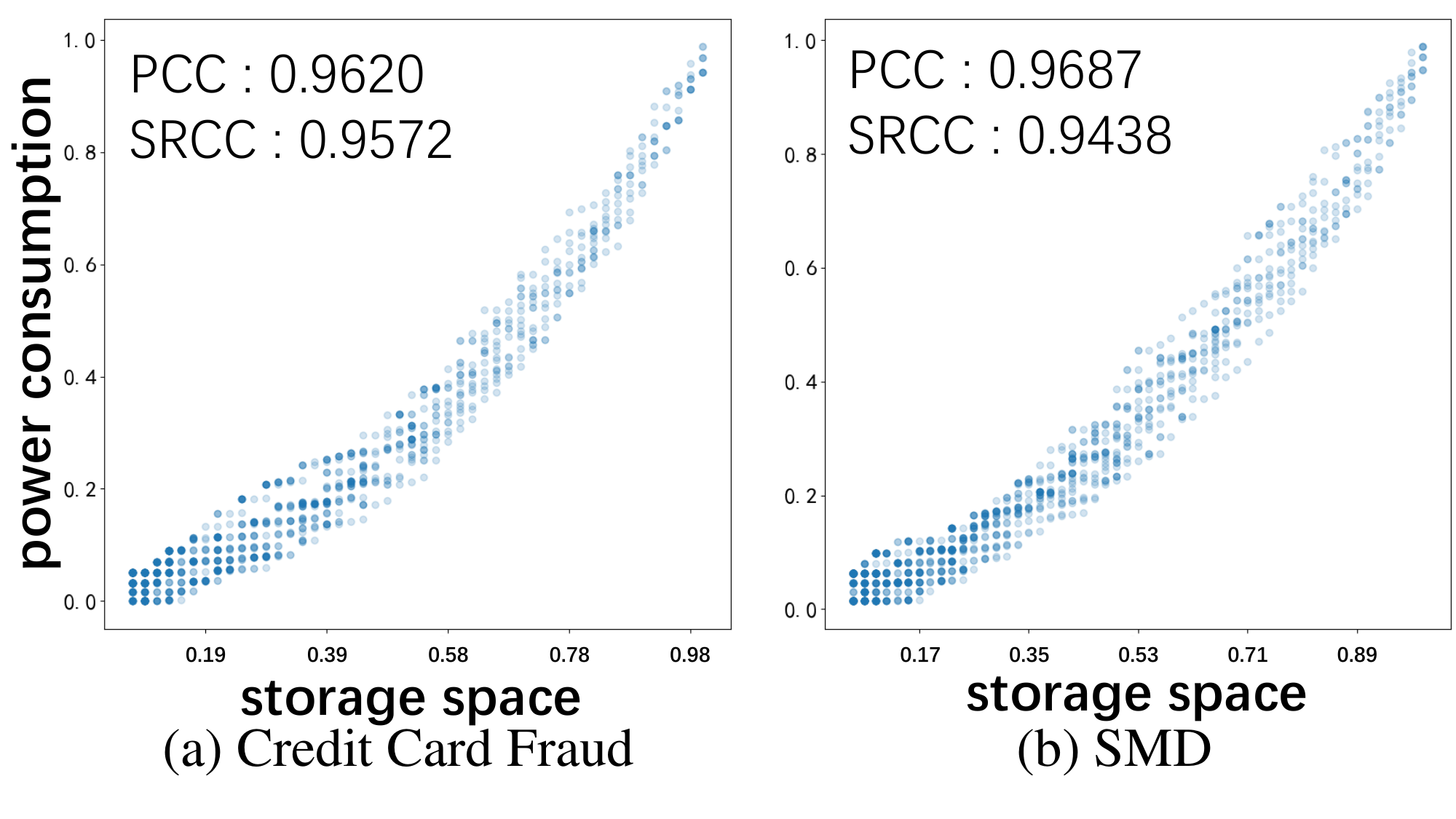}
        % \caption{Relationship between power consumption and storage space.}
        \caption{Power consumption-Storage space Relationship.}
        \label{fig:7}
        \centering
        \includegraphics[width=0.46\textwidth]{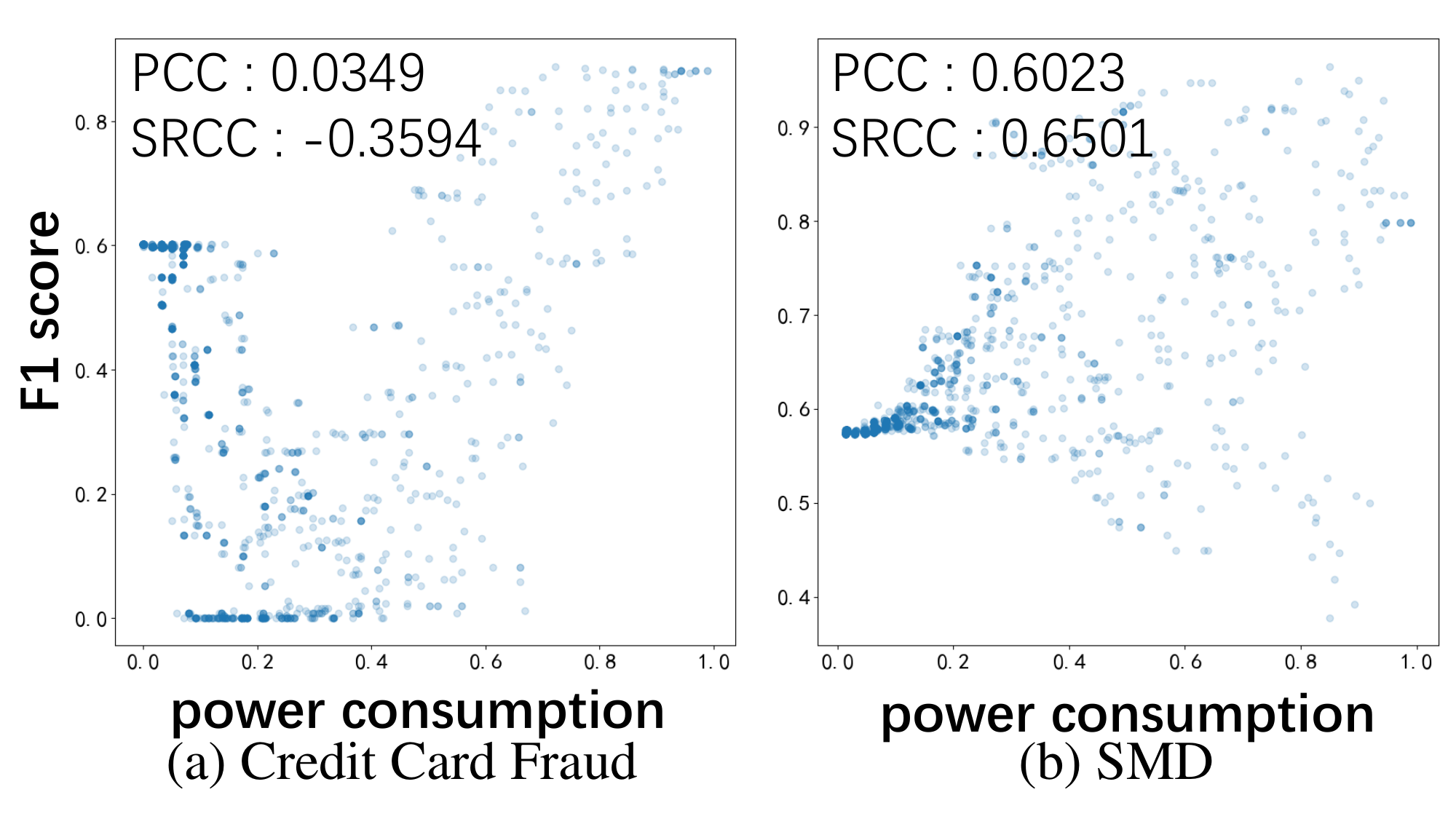}
        % \caption{Relationship between power consumption and F1 score.}
        \caption{Power consumption-F1 score Relationship.}
        \label{fig:8}
        \centering
        \includegraphics[width=0.46\textwidth]{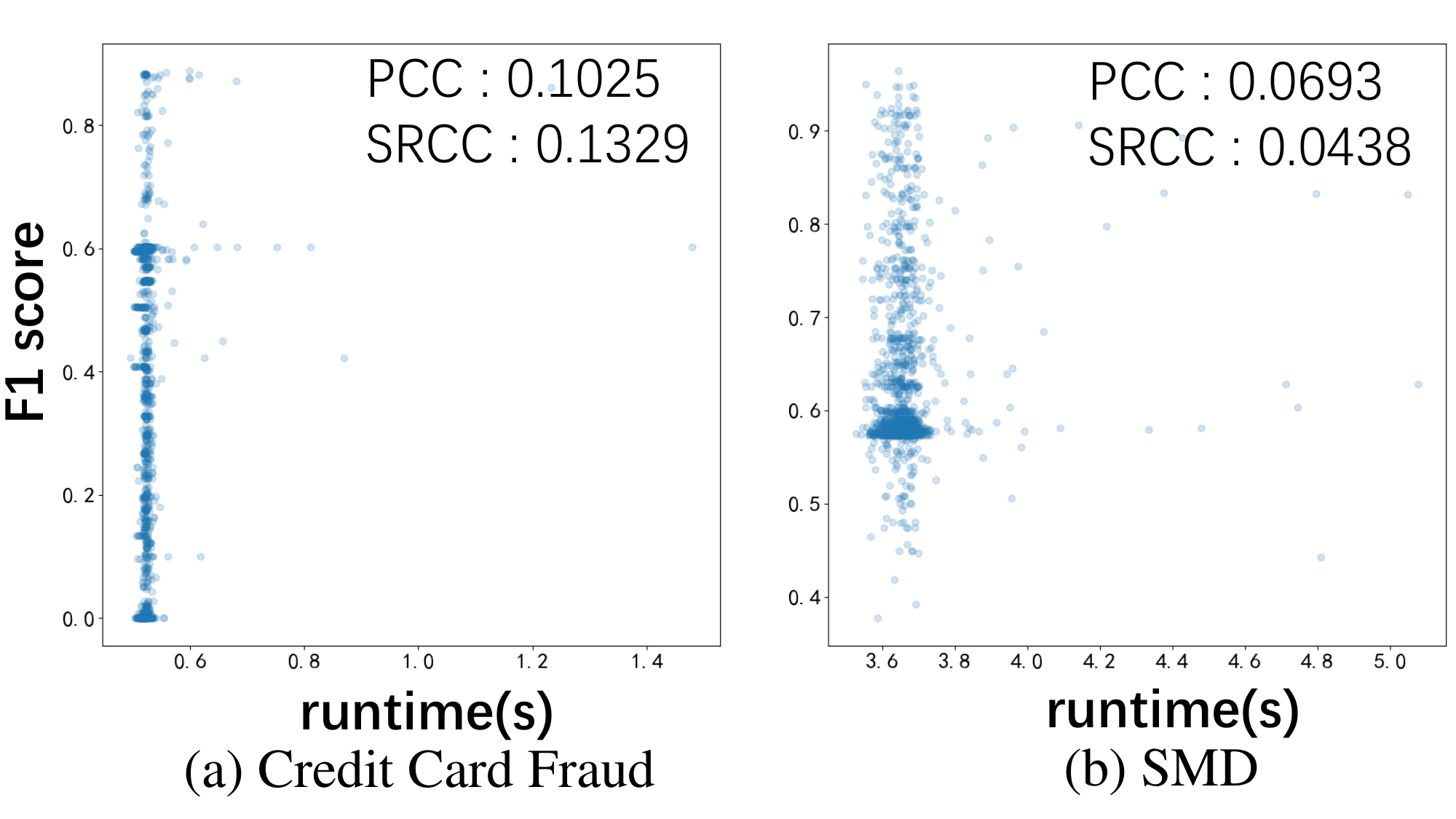}
        % \caption{Relationship between runtime and F1 score.}
        \caption{Runtime-F1 score Relationship.}
        \label{fig:9}
\end{figure}

On one hand, we aim to investigate the robustness of SAE in edge anomaly detection against structural perturbations caused by the model clipping. On the other hand, we explore the potential of optimizing SAE through the same model clipping technique and the previously introduced multi-branch exit design. To this end, we conducted experiments using the progressive clipping test and the Random Reconstruction Error Threshold (RRET) multi-branch exit design, respectively. In particular, the RRET multi-branch exit design sets exit thresholds for each branch by analyzing its reconstruction error distribution and selecting thresholds based on randomly chosen quantiles. Based on these thresholds, it makes exit decisions on a per-sample basis. In contrast, the progressive clipping test is described in Section ~\ref{Outline of Integrated Optimization Approach}. The experimental results are visualized using scatter plots, along with linear correlation analyses based on the Pearson Correlation Coefficient (PCC) and Spearman Rank Correlation Coefficient (SRCC), to illustrate the relationships among different optimization objectives.

Under the progressive clipping test, the relationships among optimization objectives exhibit varying patterns. As illustrated in Fig.~\ref{fig:7}, there is a strong linear correlation between storage space and power consumption. This aligns with intuition: More neurons required for SAE inference lead to higher computational load and, consequently, increased energy consumption. This suggests that the optimization objectives for storage space and power consumption are aligned under model clipping. The scatter data are relatively dispersed, and the linear correlation between power consumption and F1 score is illustrated in Fig.~\ref{fig:8}. Notably, the scatter plot on the Credit Card Fraud dataset exhibits a segmented V-shaped correlation. This indicates a trade-off between the optimization objectives of achieving a high F1 score and minimizing power consumption. As illustrated in Fig.~\ref{fig:9}, the runtime remains within a fixed range and exhibits little variation with F1 score, indicating that model clipping is not effective in optimizing runtime efficiency.

% \begin{figure}[ht]
%         \centering
%         \includegraphics[width=0.45\textwidth]{fig7.pdf}
%         \caption{Relationship between power consumption and storage space.}
%         \label{fig:7}
%         \centering
%         \includegraphics[width=0.45\textwidth]{fig8.pdf}
%         \caption{Relationship between power consumption and F1 score.}
%         \label{fig:8}
%         \centering
%         \includegraphics[width=0.45\textwidth]{fig9.pdf}
%         \caption{Relationship between runtime and F1 score.}
%         \label{fig:9}
% \end{figure}
% \begin{figure}[ht]
%         \centering
%         \includegraphics[width=0.45\textwidth]{fig8.pdf}
%         \caption{Relationship between power consumption and F1 score.}
%         \label{fig:8}
% \end{figure}
% \begin{figure}[ht]
%         \centering
%         \includegraphics[width=0.45\textwidth]{fig9.pdf}
%         \caption{Relationship between runtime and F1 score.}
%         \label{fig:9}
% \end{figure}

\begin{figure}[ht]
        \centering
        \includegraphics[width=0.46\textwidth]{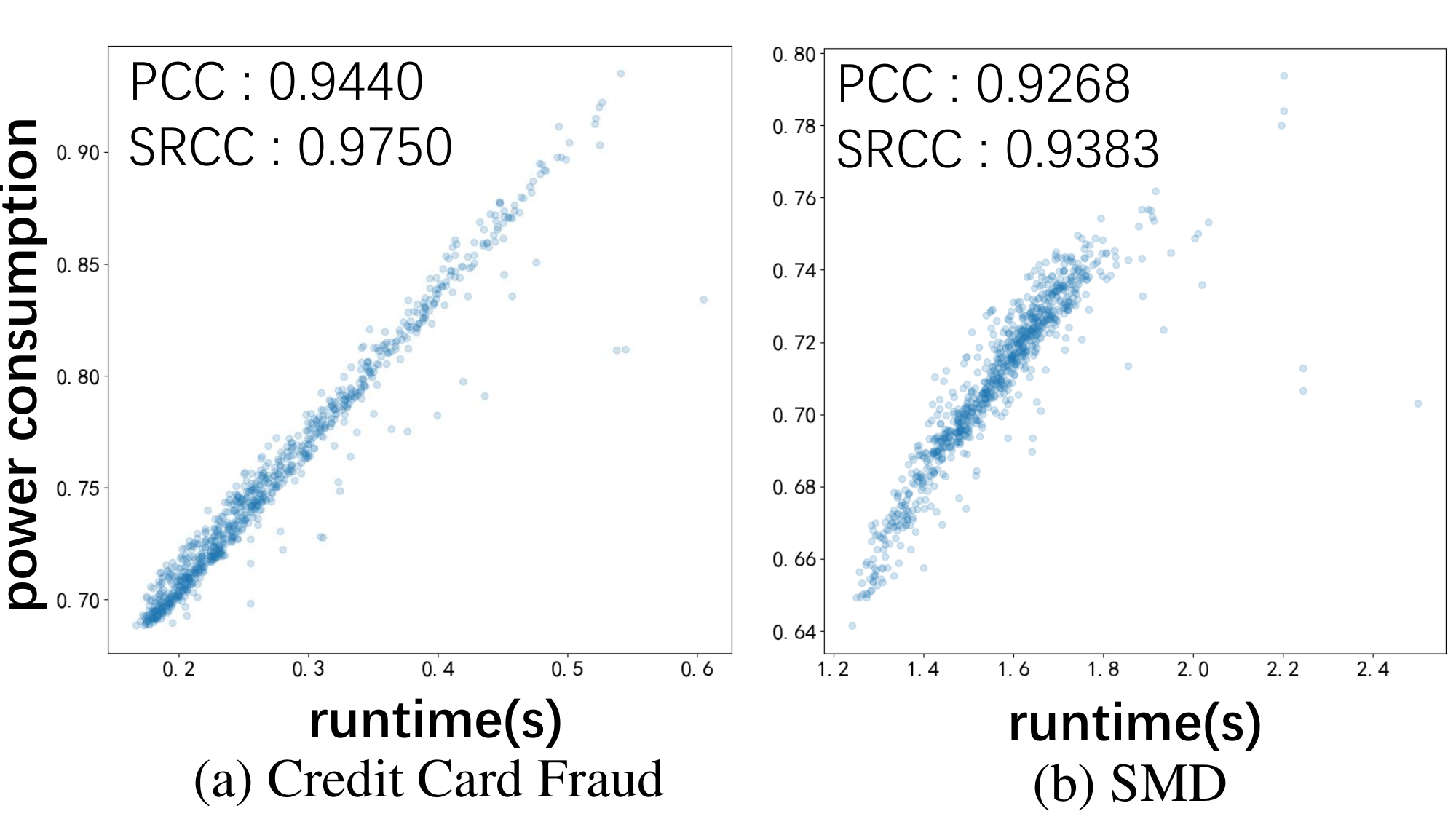}
        % \caption{Relationship between runtime and power consumption.}
        \caption{Runtime-Power consumption Relationship.}
        \label{fig:10}
        \centering
        \includegraphics[width=0.46\textwidth]{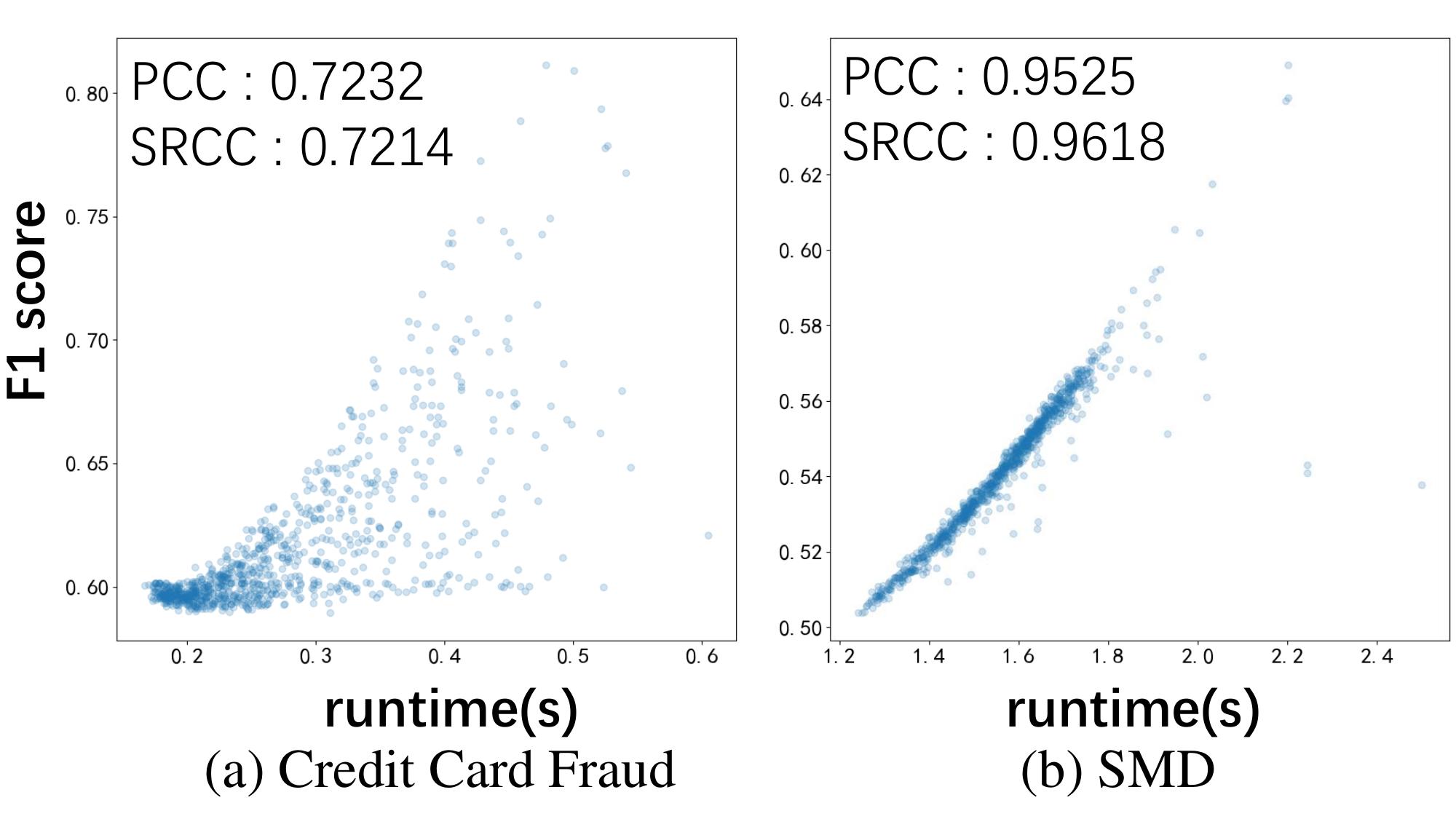}
        % \caption{Relationship between runtime and F1 score.}
        \caption{Runtime-F1 score Relationship.}
        \label{fig:11}
        \centering
        \includegraphics[width=0.46\textwidth]{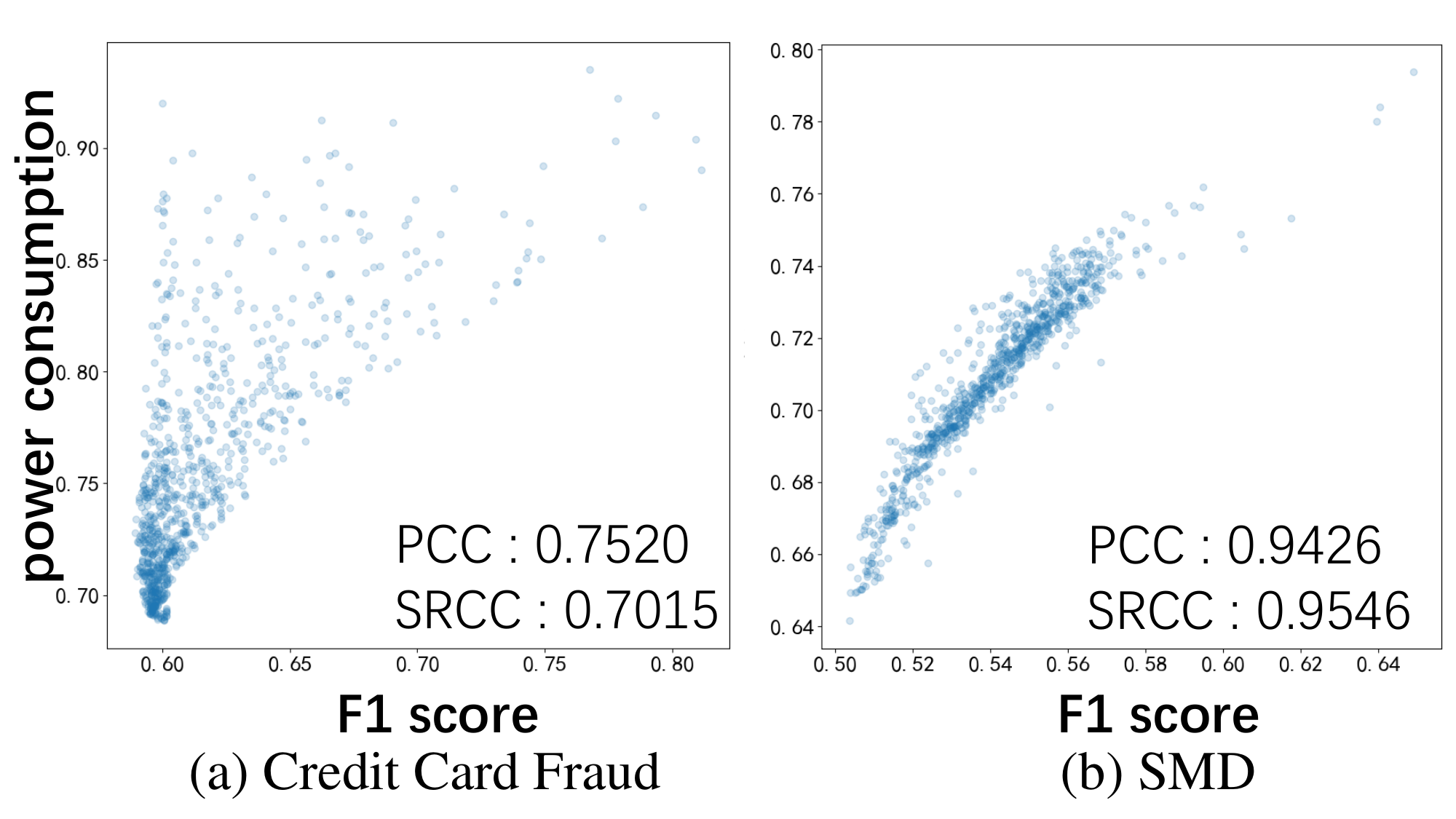}
        % \caption{Relationship between F1 score and power consumption.}
        \caption{F1 score-Power consumption Relationship.}
        \label{fig:12}
\end{figure}

The analysis of Figs. ~\ref{fig:10}, ~\ref{fig:11}, and ~\ref{fig:12} reveals that under the RRET multi-branch exit design, there exists a correlation between any two of the following objectives: runtime, power consumption, and F1 score. Given the study’s optimization goals—minimizing runtime and power consumption while maximizing F1 score—it is evident that significant trade-offs exist among these objectives. Another noteworthy observation is that the wide variation in F1 score suggests there is still room for improvement. The multi-branch exit design also proves effective in reducing runtime, thereby improving overall execution efficiency.

% \begin{figure}[ht]
%         \centering
%         \includegraphics[width=0.45\textwidth]{fig11.pdf}
%         \caption{Relationship between runtime and F1 score.}
%         \label{fig:11}
% \end{figure}
% \begin{figure}[ht]
%         \centering
%         \includegraphics[width=0.45\textwidth]{fig12.pdf}
%         \caption{Relationship between F1 score and power consumption.}
%         \label{fig:12}
% \end{figure}

\begin{figure}[ht]
        \centering
        \includegraphics[width=0.46\textwidth]{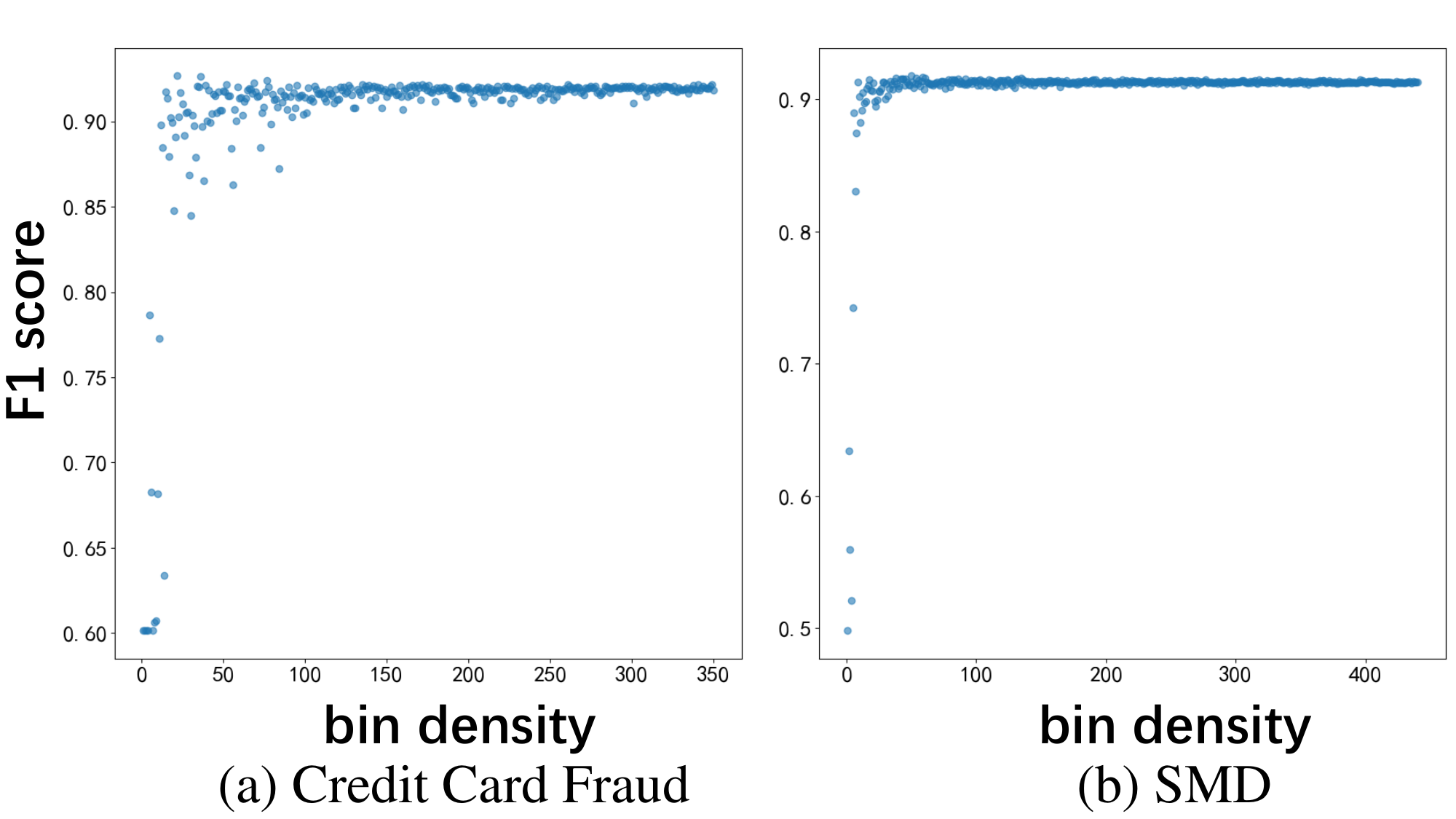}
        \caption{Variation of F1 score with Increasing bin density.}
        \label{fig:13}
\end{figure}
\begin{figure}[ht]
        \centering
        \includegraphics[width=0.46\textwidth]{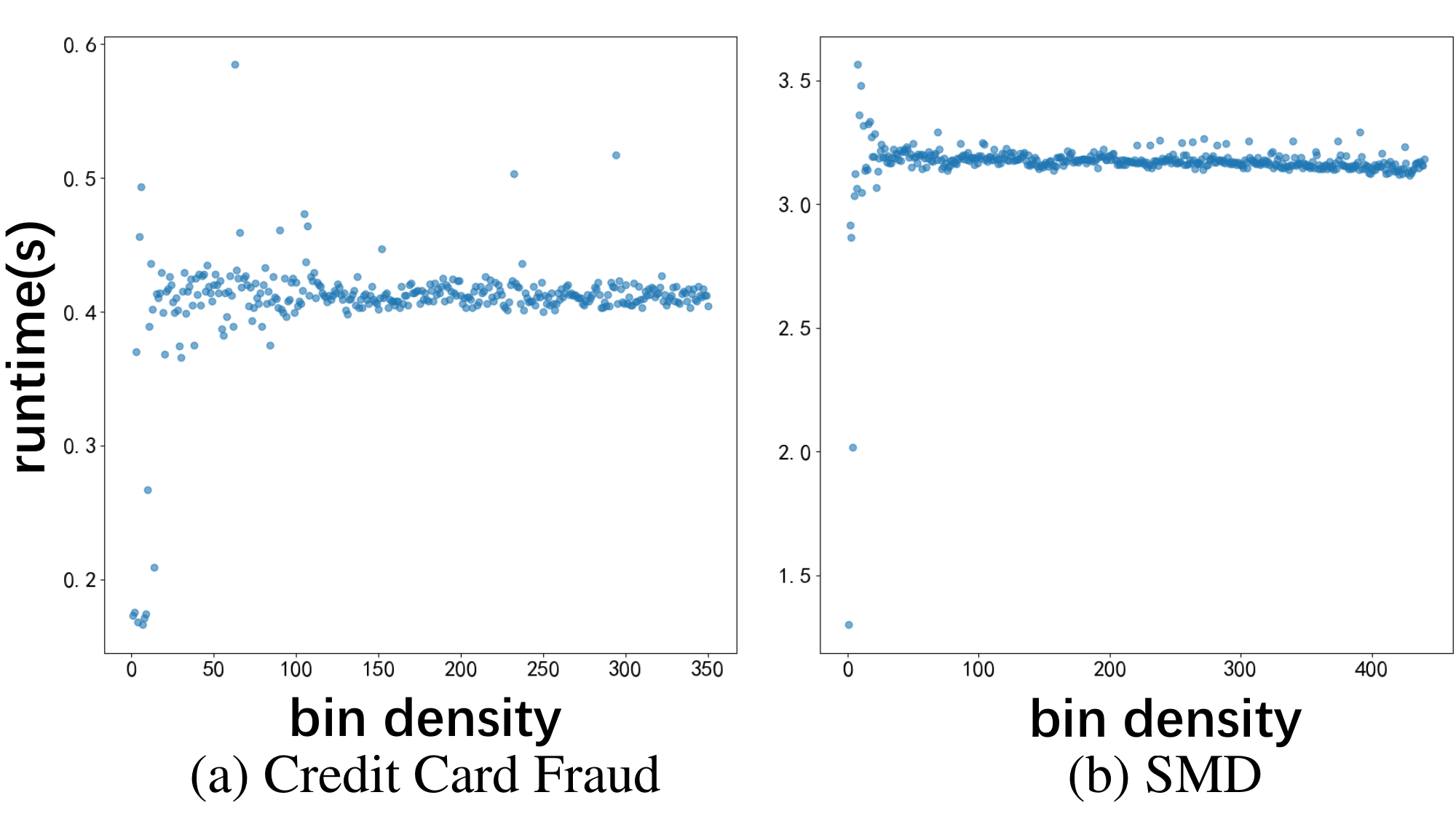}
        \caption{Variation of runtime with Increasing bin density.}
        \label{fig:14}
\end{figure}

\subsection{Effective Trade-offs by Multi-objective Heuristic Algorithm} \label{Effective Trade-offs by Multi-objective Heuristic Algorithm}

Quantitative experiments on both datasets present the results for the original SAE, MO-SAE without the model clipping, MO-SAE without the multi-branch exit design, and MO-SAE with both optimizations. 
% These results are used to analyze and compare performance with other methods. 
The baseline models for Credit Card Fraud dataset include GRU, LSTM, SVM, KNN, ANN, and LSTM-Attention. The major reason for selecting these models is that they cover a wide range of machine learning paradigms, including traditional classification (SVM, KNN), shallow neural networks (ANN), recurrent neural networks (LSTM, GRU), and attention-based sequence models (LSTM-Attention). These models have demonstrated competitive performance in previous fraud detection research and are representative of different levels of model complexity and temporal modeling capabilities. 
DAGMM, LSTM-VAE, and OmniAnomaly are selected as representative baseline models for comparison on the SMD dataset, as they cover a spectrum of modeling paradigms—from static density estimation to temporal generative modeling—and allow for comprehensive evaluation of anomaly detection performance in multivariate time series under edge environments.
As illustrated in Table~\ref{tab:1} and Table~\ref{tab:2}, bold font is used to indicate the top two values for each optimization objective, while underlines denote unacceptable F1 scores. MO-SAE employing a single optimization method—such as model clipping or multi-branch exit design—achieves an acceptable F1 score on one dataset but performs unsatisfactorily on the other. In contrast, when both optimizations are applied, MO-SAE not only maintains an acceptable F1 score on both datasets, but also achieves near-optimal performance in terms of storage space, power consumption, and runtime efficiency. 
% The experimental analysis demonstrates that combining model clipping with the multi-branch exit design enables MO-SAE to achieve significant performance improvements and better generalization across diverse scenarios. 
The experimental analysis demonstrates that the combination of model clipping and the multi-branch exit design enables MO-SAE to achieve superior generalization across diverse scenarios, compared to using either model clipping or the multi-branch exit design alone. Overall, this joint approach ranks among the top two across key performance metrics.
Furthermore, as illustrated in Table~\ref{tab:3} and Table~\ref{tab:4}, validation on a single edge device under diverse ARM configurations confirms that the combination of model clipping and the multi-branch exit design allows MO-SAE to achieve robust performance while maintaining high efficiency, minimizing storage space and power consumption across both x86 and ARM platforms.

\begin{table*}[ht]
\centering
\caption{Performance comparison of different optimized schemes on Credit Card Fraud dataset.}
\label{tab:1}
\begin{tabularx}{\linewidth}{@{}>{\raggedright\arraybackslash}X
                                >{\centering\arraybackslash}p{2.5cm}
                                >{\centering\arraybackslash}p{2.5cm}
                                >{\centering\arraybackslash}p{2.5cm}
                                >{\centering\arraybackslash}p{3.6cm}@{}}
\toprule
\textbf{Methods} & \textbf{F1 score} & \textbf{Storage space ratio} & \textbf{Runtime (s)} & \textbf{Power Consumption ratio} \\
\midrule
GRU \cite{credit-card-fraud-compare}             & 0.78              & --                          & --                  & -- \\  % ← 新增行
LSTM \cite{credit-card-fraud-compare}            & 0.79              & 12.35                        & 21.90                  & 49.46 \\  % ← 新增行
SVM \cite{credit-card-fraud-compare}             & 0.93              & --                          & --                  & -- \\  % ← 新增行
KNN \cite{credit-card-fraud-compare}             & 0.07              & --                          & --                  & -- \\  % ← 新增行
ANN \cite{credit-card-fraud-compare}             & 0.78              & --                          & --                  & -- \\  % ← 新增行
LSTM-attention \cite{credit-card-fraud-compare}      & \textbf{0.95}              & 12.37                        & 19.35                  & 49.55 \\  % ← 新增行
Original                   & 0.88              & 1.00                        & 0.70                & 1.00                              \\
MO-SAE without Model clipping   & 0.85              & 1.00                        & \textbf{0.48}                & 0.92                              \\
MO-SAE without Multi-branch exit design             & \underline{0.60}              & \textbf{0.10}                        & 0.57                & \textbf{0.04}                              \\
MO-SAE                     & \textbf{0.91}              & \textbf{0.31}                        & \textbf{0.44}                & \textbf{0.18}                              \\
\bottomrule
\end{tabularx}
\end{table*}

\begin{table*}[ht]
\centering
\caption{Performance comparison of different optimized schemes on SMD dataset.}
\label{tab:2}
\begin{tabularx}{\linewidth}{@{}>{\raggedright\arraybackslash}X
                                >{\centering\arraybackslash}p{2.5cm}
                                >{\centering\arraybackslash}p{2.5cm}
                                >{\centering\arraybackslash}p{2.5cm}
                                >{\centering\arraybackslash}p{3.6cm}@{}}
\toprule
\textbf{Methods} & \textbf{F1 score} & \textbf{Storage space ratio} & \textbf{Runtime (s)} & \textbf{Power Consumption ratio} \\
\midrule
DAGMM \cite{OmniAnomaly}           & 0.70              & --                          & --                  & -- \\  % ← 新增行
LSTM-VAE \cite{OmniAnomaly}        & 0.78              & --                          & --                  & -- \\  % ← 新增行
OmniAnomaly \cite{OmniAnomaly}     & 0.88              & 831.01                      & 128.63              & 1093069.97 \\  % ← 新增行
Original                   & 0.79              & 1.00                        & 4.88                & 1.00                              \\
MO-SAE without Model clipping   & \underline{0.63}              & 1.00                        & \textbf{2.05}                & 0.77                              \\
MO-SAE without Multi-branch exit design             & \textbf{0.95}              & \textbf{0.45}                        & 3.92                & \textbf{0.49}                              \\
MO-SAE                     & \textbf{0.92}              & \textbf{0.47}                        & \textbf{3.49}                & \textbf{0.37}                              \\
\bottomrule
\end{tabularx}
\end{table*}

\begin{table*}[ht]
\centering
\caption{Performance comparison on Credit Card Fraud dataset under real resource-limited intelligent platform.}
\label{tab:3}
\begin{tabularx}{\textwidth}{@{}>{\raggedright\arraybackslash}m{5cm}
                                >{\centering\arraybackslash}m{1.8cm}
                                >{\centering\arraybackslash}X
                                >{\centering\arraybackslash}X
                                >{\centering\arraybackslash}X
                                >{\centering\arraybackslash}X
                                >{\centering\arraybackslash}X@{}}
\toprule
\multirow{2}{*}{\textbf{Method}} &
\multirow{2}{*}{\textbf{F1 Score}} &
\multicolumn{5}{c}{\textbf{Runtime (s) under different Work modes}} \\
\cmidrule(l){3-7}
& & \textbf{15w, 2cores} & \textbf{15w, 4cores} & \textbf{15w, 6cores} & \textbf{10w, 2cores} & \textbf{10w, 4cores} \\
\midrule
Original           & \textbf{0.88} & 1.60 & 2.03 & 2.07 & 1.88 & 2.35 \\
MO-SAE without Model clipping  & 0.85 & \textbf{1.34} & \textbf{1.76} & \textbf{1.73} & \textbf{1.65} & \textbf{2.07} \\
MO-SAE without Multi-branch exit design     & \underline{0.60} & 1.57 & 2.00 & 2.05 & 1.83 & 2.36 \\
MO-SAE             & \textbf{0.91} & \textbf{1.16} & \textbf{1.54} & \textbf{1.52} & \textbf{1.45} & \textbf{1.83} \\
\bottomrule
\end{tabularx}
\end{table*}

\begin{table*}[ht]
\centering
\caption{Performance comparison on SMD dataset under real resource-limited intelligent platform.}
\label{tab:4}
\begin{tabularx}{\textwidth}{@{}>{\raggedright\arraybackslash}m{5cm}
                                >{\centering\arraybackslash}m{1.8cm}
                                >{\centering\arraybackslash}X
                                >{\centering\arraybackslash}X
                                >{\centering\arraybackslash}X
                                >{\centering\arraybackslash}X
                                >{\centering\arraybackslash}X@{}}
\toprule
\multirow{2}{*}{\textbf{Method}} &
\multirow{2}{*}{\textbf{F1 Score}} &
\multicolumn{5}{c}{\textbf{Runtime (s) under different Work modes}} \\
\cmidrule(l){3-7}
& & \textbf{15w, 2cores} & \textbf{15w, 4cores} & \textbf{15w, 6cores} & \textbf{10w, 2cores} & \textbf{10w, 4cores} \\
\midrule
Original           & 0.79 & 10.29 & 13.73 & 13.86 & 13.29 & 16.59 \\
MO-SAE without Model clipping  & \underline{0.63} & \textbf{5.58} & \textbf{7.44} & \textbf{7.47} & \textbf{7.15} & \textbf{9.01} \\
MO-SAE without Multi-branch exit design     & \textbf{0.95} & 10.21 & 13.67 & 13.73 & 13.06 & 16.48 \\
MO-SAE             & \textbf{0.92} & \textbf{8.87} & \textbf{11.88} & \textbf{11.81} & \textbf{11.33} & \textbf{13.28} \\
\bottomrule
\end{tabularx}
\end{table*}

\subsection{Evaluation of Model Update Optimization}

As illustrated in Fig.~\ref{fig:14}, the key to model update optimization via equal-width binning lies in selecting an appropriate bin density. A well-chosen bin density has a negligible impact on the model’s runtime, ensuring that the runtime optimization benefits discussed in ~\ref{Effective Trade-offs by Multi-objective Heuristic Algorithm} are not compromised. As illustrated in Fig.~\ref{fig:13}, the F1 score increases gradually with the rising bin density and eventually reaches a plateau. An overly small bin density can lead to a significant decline in model performance, while an excessively large bin density may result in unnecessary bandwidth waste and model update inefficiency, especially in cloud-edge collaboration scenarios. To ensure MO-SAE delivers stable performance, this study sets bin density to 100 for both datasets, achieving an 11.08\% compression rate compared to the original.
%To ensure edge anomaly detection SAE delivers stable performance, this study sets bins density to 100 for the CreditCardFraud and SMD datasets, achieving a 11.08\% compression rate compared to the state before model update optimization.

% \begin{figure}[ht]
%         \centering
%         \includegraphics[width=0.45\textwidth]{fig13.pdf}
%         \caption{Variation of F1 score with Increasing bin density.}
%         \label{fig:13}
%         \centering
%         \includegraphics[width=0.45\textwidth]{fig14.pdf}
%         \caption{Variation of runtime with Increasing bin density.}
%         \label{fig:14}
% \end{figure}

% \begin{figure}[ht]
%         \centering
%         \includegraphics[width=0.45\textwidth]{fig14.pdf}
%         \caption{Variation of runtime with Increasing bins density.}
%         \label{fig:14}
% \end{figure}

\section{CONCLUSIONS}

In this paper, we propose an integrated optimization framework that systematically optimizes SAE from multiple aspects, including application performance, storage space, power consumption, runtime efficiency, and model update efficiency. Based on model clipping and a multi-branch exit design, we introduce a multi-objective heuristic algorithm to balance trade-offs among competing optimization objectives. Additionally, grounded in solid theoretical analysis and quantitative experiments results, we develop model update optimization tailored for the transmission process. Extensive experiments on real-world datasets validate the effectiveness of MO-SAE. 

%First, SAE in edge anomaly detection exhibit strong robustness to structural perturbations by model clipping and possess significant untapped optimization potential. Second, our approach demonstrates substantial strengths by simultaneously considering multiple factors in edge anomaly detection scenarios. Finally, based on theoretical analysis and comprehensive evaluation, the effective model update optimization will contribute to edge anomaly detection SAE deployment across various scenarios.


%%%%%%%%%%%%%%%%%%%%%%%%%%%%%%%%%%%%%%%%%%%%%%%%%%%%%%%%%%%%%%%%%%%%%%%%%%%%%%%%

%%%%%%%%%%%%%%%%%%%%%%%%%%%%%%%%%%%%%%%%%%%%%%%%%%%%%%%%%%%%%%%%%%%%%%%%%%%%%%%%



\section*{ACKNOWLEDGMENT}

% The preferred spelling of the word ÒacknowledgmentÓ in America is without an ÒeÓ after the ÒgÓ. Avoid the stilted expression, ÒOne of us (R. B. G.) thanks . . .Ó  Instead, try ÒR. B. G. thanksÓ. Put sponsor acknowledgments in the unnumbered footnote on the first page.
This work was supported by National Natural Science Foundation of China under Grant No. 62406087, Shandong Natural Science Foundation of China under Grant No.  ZR2024QF139, State Key Laboratory of Processors (ICT, CAS) under Grant No. CLQ202406, and State Key Laboratory of Computer Architecture (ICT,CAS) under Grant No. CARCHA202104.

%%%%%%%%%%%%%%%%%%%%%%%%%%%%%%%%%%%%%%%%%%%%%%%%%%%%%%%%%%%%%%%%%%%%%%%%%%%%%%%%

% References are important to the reader; therefore, each citation must be complete and correct. If at all possible, references should be commonly available publications.


\begin{thebibliography}{99}

% \bibitem{SAE-anomaly-detect} D. Liu, Y. Wang, C. Liu, X. Yuan, C. Yang, \& W. Gui, Data mode related interpretable transformer network for predictive modeling and key sample analysis in industrial processes, {\em IEEE Trans. Ind. Inf.}, \textbf{19}(9), 9325--9336, Sep. 2023. {https://doi.org/10.1109/TII.2022.3227731}
\bibitem{SAE-anomaly-detect}
D. Liu, Y. Wang, C. Liu, X. Yuan, C. Yang, and W. Gui,  
"Data mode related interpretable transformer network for predictive modeling and key sample analysis in industrial processes,"  
\emph{IEEE Trans. Ind. Inf.}, vol. 19, no. 9, pp. 9325--9336, Sep. 2023.  
doi: \href{https://doi.org/10.1109/TII.2022.3227731}{10.1109/TII.2022.3227731}

% \bibitem{resource-constraints} L. Liu, G. G. Yen, \& Z. He, \emph{EvolutionViT: Multi-objective evolutionary vision transformer pruning under resource constraints}, \emph{Information Sciences}, \textbf{689}, 121406, Jan. 2025. {https://doi.org/10.1016/j.ins.2024.121406}
\bibitem{resource-constraints}
L. Liu, G. G. Yen, and Z. He,  
"EvolutionViT: Multi-objective evolutionary vision transformer pruning under resource constraints,"  
\emph{Information Sciences}, vol. 689, Art. no. 121406, Jan. 2025.  
doi: \href{https://doi.org/10.1016/j.ins.2024.121406}{10.1016/j.ins.2024.121406}

% \bibitem{DNN-acclerate-survey} Cheng, H., Zhang, M. \& Shi, J. A survey on deep neural network pruning: Taxonomy, comparison, analysis, and recommendations. {\em IEEE Transactions On Pattern Analysis And Machine Intelligence}. (2024)
\bibitem{DNN-acclerate-survey}
H. Cheng, M. Zhang, and J. Shi,  
"A survey on deep neural network pruning: Taxonomy, comparison, analysis, and recommendations,"  
\emph{IEEE Trans. Pattern Anal. Mach. Intell.}, 2024, early access.

% \bibitem{KD-anomaly-detect} Thomine, S. \& Snoussi, H. Dual model knowledge distillation for industrial anomaly detection. {\em Pattern Anal. Appl.}. \textbf{27}, 77 (2024), https://doi.org/10.1007/s10044-024-01295-8
\bibitem{KD-anomaly-detect}
S. Thomine and H. Snoussi,  
"Dual model knowledge distillation for industrial anomaly detection,"  
\emph{Pattern Anal. Appl.}, vol. 27, Art. no. 77, 2024.  
doi: \href{https://doi.org/10.1007/s10044-024-01295-8}{10.1007/s10044-024-01295-8}

% \bibitem{knowledge-distillation-survey} S. Li, T. Su, X.-Y. Zhang, and Z. Wang, "Continual learning with knowledge distillation: A survey," IEEE Trans. Neural Netw. Learn. Syst., vol. 35, no. 1, pp. 1–21, 2024. doi: 10.1109/TNNLS.2024.3476068.
\bibitem{knowledge-distillation-survey}
S. Li, T. Su, X.-Y. Zhang, and Z. Wang,  
"Continual learning with knowledge distillation: A survey,"  
\emph{IEEE Trans. Neural Netw. Learn. Syst.}, vol. 35, no. 1, pp. 1--21, 2024.  
doi: \href{https://doi.org/10.1109/TNNLS.2024.3476068}{10.1109/TNNLS.2024.3476068}

% \bibitem{low-rank} A. Novikov, D. Podoprikhin, A. Osokin, \& D. P. Vetrov, \emph{Tensorizing neural networks}, in \emph{Proceedings of the 29th International Conference on Neural Information Processing Systems (NeurIPS)}, Montreal, Canada, 2015, pp. 442--450.
\bibitem{low-rank}
A. Novikov, D. Podoprikhin, A. Osokin, and D. P. Vetrov,  
"Tensorizing neural networks,"  
in \emph{Proc. 29th Int. Conf. Neural Inf. Process. Syst. (NeurIPS)}, Montreal, Canada, 2015, pp. 442--450.

% \bibitem{mathematical-conditions} P. Wang \& J. Cheng, \emph{Accelerating convolutional neural networks for mobile applications}, in \emph{Proceedings of the 24th ACM International Conference on Multimedia (ACM MM)}, Amsterdam, The Netherlands, Oct. 2016, pp. 541--545. {https://doi.org/10.1145/2964284.2967280}
\bibitem{mathematical-conditions}
P. Wang and J. Cheng,  
"Accelerating convolutional neural networks for mobile applications,"  
in \emph{Proc. 24th ACM Int. Conf. Multimedia (ACM MM)}, Amsterdam, The Netherlands, Oct. 2016, pp. 541--545.  
doi: \href{https://doi.org/10.1145/2964284.2967280}{10.1145/2964284.2967280}

% \bibitem{Para-meter-sharing} W. Chen, J. Wilson, S. Tyree, K. Q. Weinberger, \& Y. Chen, \emph{Compressing convolutional neural networks in the frequency domain}, in \emph{Proceedings of the 22nd ACM SIGKDD International Conference on Knowledge Discovery and Data Mining (KDD)}, San Francisco, CA, USA, Aug. 2016, pp. 1475--1484. {https://doi.org/10.1145/2939672.2939839}
\bibitem{Parameter-sharing}
W. Chen, J. Wilson, S. Tyree, K. Q. Weinberger, and Y. Chen,  
"Compressing convolutional neural networks in the frequency domain,"  
in \emph{Proc. 22nd ACM SIGKDD Int. Conf. Knowl. Discov. Data Min. (KDD)}, San Francisco, CA, USA, Aug. 2016, pp. 1475--1484.  
doi: \href{https://doi.org/10.1145/2939672.2939839}{10.1145/2939672.2939839}

% \bibitem{Cluster} J. Wu, Y. Wang, Z. Wu, Z. Wang, A. Veeraraghavan, \& Y. Lin, \emph{Deep $k$-means: Re-training and parameter sharing with harder cluster assignments for compressing deep convolutions}, arXiv preprint arXiv:1806.09228, Jun. 2018. {https://doi.org/10.48550/arXiv.1806.09228}
\bibitem{Cluster}
J. Wu, Y. Wang, Z. Wu, Z. Wang, A. Veeraraghavan, and Y. Lin,  
"Deep $k$-means: Re-training and parameter sharing with harder cluster assignments for compressing deep convolutions,"  
\emph{arXiv preprint} arXiv:1806.09228, Jun. 2018.  
doi: \href{https://doi.org/10.48550/arXiv.1806.09228}{10.48550/arXiv.1806.09228}

% \bibitem{Hashing} W. Chen, J. T. Wilson, S. Tyree, K. Q. Weinberger, \& Y. Chen, \emph{Compressing neural networks with the hashing trick}, in \emph{Proceedings of the 32nd International Conference on Machine Learning (ICML)}, Lille, France, Jul. 2015, pp. 2285--2294.
\bibitem{Hashing}
W. Chen, J. T. Wilson, S. Tyree, K. Q. Weinberger, and Y. Chen,  
"Compressing neural networks with the hashing trick,"  
in \emph{Proc. 32nd Int. Conf. Mach. Learn. (ICML)}, Lille, France, Jul. 2015, pp. 2285--2294.

% \bibitem{theoretical-analysis} J. Frankle \& M. Carbin, \emph{The lottery ticket hypothesis: Finding sparse, trainable neural networks}, in \emph{Proceedings of the 7th International Conference on Learning Representations (ICLR)}, May 2019. {https://openreview.net/forum?id=rJl-b3RcF7}
\bibitem{theoretical-analysis}
J. Frankle and M. Carbin,  
"The lottery ticket hypothesis: Finding sparse, trainable neural networks,"  
in \emph{Proc. 7th Int. Conf. Learn. Represent. (ICLR)}, May 2019.  
Available: \href{https://openreview.net/forum?id=rJl-b3RcF7}{https://openreview.net/forum?id=rJl-b3RcF7}

% \bibitem{Parameter-quantization} D. Soudry, I. Hubara, \& R. Meir, \emph{Expectation backpropagation: Parameter-free training of multilayer neural networks with continuous or discrete weights}, in \emph{Proceedings of the 28th International Conference on Neural Information Processing Systems (NeurIPS)}, Montreal, Canada, Dec. 2014, pp. 963--971.
\bibitem{Parameter-quantization}
D. Soudry, I. Hubara, and R. Meir,  
"Expectation backpropagation: Parameter-free training of multilayer neural networks with continuous or discrete weights,"  
in \emph{Proc. 28th Int. Conf. Neural Inf. Process. Syst. (NeurIPS)}, Montreal, Canada, Dec. 2014, pp. 963--971.

% \bibitem{Depth-Pruning} F. Yu, K. Huang, M. Wang, Y. Cheng, W. Chu, \& L. Cui, \emph{Width \& depth pruning for vision transformers}, \emph{Proceedings of the AAAI Conference on Artificial Intelligence}, \textbf{36}(3), pp. 3143--3151, Jun. 2022. {https://doi.org/10.1609/aaai.v36i3.20222}
\bibitem{Depth-Pruning}
F. Yu, K. Huang, M. Wang, Y. Cheng, W. Chu, and L. Cui,  
"Width \& depth pruning for vision transformers,"  
in \emph{Proc. AAAI Conf. Artif. Intell.}, vol. 36, no. 3, pp. 3143--3151, Jun. 2022.  
doi: \href{https://doi.org/10.1609/aaai.v36i3.20222}{10.1609/aaai.v36i3.20222}

% \bibitem{kernel-level} S. Lin, R. Ji, Y. Li, C. Deng, \& X. Li, \emph{Towards compact ConvNets via structure-sparsity regularized filter pruning}, arXiv preprint arXiv:1901.07827, Mar. 2019. {https://doi.org/10.48550/arXiv.1901.07827}
\bibitem{kernel-level}
S. Lin, R. Ji, Y. Li, C. Deng, and X. Li,  
"Towards compact ConvNets via structure-sparsity regularized filter pruning,"  
\emph{arXiv preprint} arXiv:1901.07827, Mar. 2019.  
doi: \href{https://doi.org/10.48550/arXiv.1901.07827}{10.48550/arXiv.1901.07827}

% \bibitem{filter-level} D. Mehta, K. I. Kim, \& C. Theobalt, \emph{On implicit filter level sparsity in convolutional neural networks}, in \emph{Proceedings of the IEEE/CVF Conference on Computer Vision and Pattern Recognition (CVPR)}, Long Beach, CA, USA, Jun. 2019, pp. 520--528. {https://doi.org/10.1109/CVPR.2019.00061}
\bibitem{filter-level}
D. Mehta, K. I. Kim, and C. Theobalt,  
"On implicit filter level sparsity in convolutional neural networks,"  
in \emph{Proc. IEEE/CVF Conf. Comput. Vis. Pattern Recognit. (CVPR)}, Long Beach, CA, USA, Jun. 2019, pp. 520--528.  
doi: \href{https://doi.org/10.1109/CVPR.2019.00061}{10.1109/CVPR.2019.00061}

% \bibitem{vector-level} S. Li, K. Osawa, \& T. Hoefler, \emph{Efficient quantized sparse matrix operations on tensor cores}, in \emph{Proceedings of the SC22: International Conference for High Performance Computing, Networking, Storage and Analysis}, Dallas, TX, USA, Nov. 2022, pp. 1--15. {https://doi.org/10.1109/SC41404.2022.00042}
\bibitem{vector-level}
S. Li, K. Osawa, and T. Hoefler,  
"Efficient quantized sparse matrix operations on tensor cores,"  
in \emph{Proc. SC22: Int. Conf. High Perform. Comput., Netw., Storage and Anal.}, Dallas, TX, USA, Nov. 2022, pp. 1--15.  
doi: \href{https://doi.org/10.1109/SC41404.2022.00042}{10.1109/SC41404.2022.00042}

% \bibitem{NSIBF} C. Feng \& P. Tian, \emph{Time series anomaly detection for cyber-physical systems via neural system identification and Bayesian filtering}, in \emph{Proceedings of the 27th ACM SIGKDD Conference on Knowledge Discovery and Data Mining (KDD)}, Virtual Event, Singapore, Aug. 2021, pp. 2858--2867. {https://doi.org/10.1145/3447548.3467137}
\bibitem{NSIBF}
C. Feng and P. Tian,  
"Time series anomaly detection for cyber-physical systems via neural system identification and Bayesian filtering,"  
in \emph{Proc. 27th ACM SIGKDD Conf. Knowl. Discov. Data Min. (KDD)}, Virtual Event, Singapore, Aug. 2021, pp. 2858--2867.  
doi: \href{https://doi.org/10.1145/3447548.3467137}{10.1145/3447548.3467137}

% \bibitem{GANF} E. Dai \& J. Chen, \emph{Graph-augmented normalizing flows for anomaly detection of multiple time series}, arXiv preprint arXiv:2202.07857, May 2022. {https://doi.org/10.48550/arXiv.2202.07857}
\bibitem{GANF}
E. Dai and J. Chen,  
"Graph-augmented normalizing flows for anomaly detection of multiple time series,"  
\emph{arXiv preprint} arXiv:2202.07857, May 2022.  
doi: \href{https://doi.org/10.48550/arXiv.2202.07857}{10.48550/arXiv.2202.07857}

% \bibitem{OmniAnomaly} Y. Su, Y. Zhao, C. Niu, R. Liu, W. Sun, \& D. Pei, \emph{Robust anomaly detection for multivariate time series through stochastic recurrent neural network}, arXiv preprint, 2019. {https://arxiv.org/abs/1802.09587}
\bibitem{OmniAnomaly}
Y. Su, Y. Zhao, C. Niu, R. Liu, W. Sun, and D. Pei,  
"Robust anomaly detection for multivariate time series through stochastic recurrent neural network,"  
\emph{arXiv preprint} arXiv:1802.09587, 2019.  
Available: \href{https://arxiv.org/abs/1802.09587}{https://arxiv.org/abs/1802.09587}

% \bibitem{LSTM} K. Hundman, V. Constantinou, C. Laporte, I. Colwell, \& T. Soderstrom, \emph{Detecting spacecraft anomalies using LSTMs and nonparametric dynamic thresholding}, in \emph{Proceedings of the 24th ACM SIGKDD International Conference on Knowledge Discovery and Data Mining (KDD)}, London, United Kingdom, Jul. 2018, pp. 387--395. {https://doi.org/10.1145/3219819.3219845}
\bibitem{LSTM}
K. Hundman, V. Constantinou, C. Laporte, I. Colwell, and T. Soderstrom,  
"Detecting spacecraft anomalies using LSTMs and nonparametric dynamic thresholding,"  
in \emph{Proc. 24th ACM SIGKDD Int. Conf. Knowl. Discov. Data Min. (KDD)}, London, U.K., Jul. 2018, pp. 387--395.  
doi: \href{https://doi.org/10.1145/3219819.3219845}{10.1145/3219819.3219845}

% \bibitem{LegoDNN-software1} R. Han, Q. Zhang, C. H. Liu, G. Wang, J. Tang, \& L. Y. Chen, \emph{LegoDNN: block-grained scaling of deep neural networks for mobile vision}, in \emph{Proceedings of the 27th Annual International Conference on Mobile Computing and Networking (MobiCom)}, New Orleans, Louisiana, Oct. 2021, pp. 406--419. {https://doi.org/10.1145/3447993.3483249}
\bibitem{LegoDNN-software1}
R. Han, Q. Zhang, C. H. Liu, G. Wang, J. Tang, and L. Y. Chen,  
"LegoDNN: Block-grained scaling of deep neural networks for mobile vision,"  
in \emph{Proc. 27th Annu. Int. Conf. Mobile Comput. Netw. (MobiCom)}, New Orleans, LA, USA, Oct. 2021, pp. 406--419.  
doi: \href{https://doi.org/10.1145/3447993.3483249}{10.1145/3447993.3483249}

% \bibitem{Design-Space-Exploration-software2} Z. Zhao, K. M. Barijough, \& A. Gerstlauer, \emph{Network-level design space exploration of resource-constrained networks-of-systems}, \emph{ACM Transactions on Embedded Computing Systems (TECS)}, vol. 19, no. 4, pp. 1--26, Jul. 2020. {https://doi.org/10.1145/3387918}
\bibitem{Design-Space-Exploration-software2}
Z. Zhao, K. M. Barijough, and A. Gerstlauer,  
"Network-level design space exploration of resource-constrained networks-of-systems,"  
\emph{ACM Trans. Embed. Comput. Syst.}, vol. 19, no. 4, pp. 1--26, Jul. 2020.  
doi: \href{https://doi.org/10.1145/3387918}{10.1145/3387918}

% \bibitem{Credit-Card-Fraud-dataset1} E. Ileberi, Y. Sun, \& Z. Wang, \emph{A machine learning based credit card fraud detection using the GA algorithm for feature selection}, \emph{Journal of Big Data}, vol. 9, no. 1, p. 24, Dec. 2022. {https://doi.org/10.1186/s40537-022-00573-8}
\bibitem{Credit-Card-Fraud-dataset1}
E. Ileberi, Y. Sun, and Z. Wang,  
"A machine learning based credit card fraud detection using the GA algorithm for feature selection,"  
\emph{J. Big Data}, vol. 9, no. 1, Art. no. 24, Dec. 2022.  
doi: \href{https://doi.org/10.1186/s40537-022-00573-8}{10.1186/s40537-022-00573-8}

% \bibitem{BranchyNet-software3} S. Teerapittayanon, B. McDanel, \& H. T. Kung, \emph{BranchyNet: fast inference via early exiting from deep neural networks}, arXiv preprint arXiv:1709.01686, Sep. 2017. {https://doi.org/10.48550/arXiv.1709.01686}
\bibitem{BranchyNet-software3}
S. Teerapittayanon, B. McDanel, and H. T. Kung,  
"BranchyNet: Fast inference via early exiting from deep neural networks,"  
\emph{arXiv preprint} arXiv:1709.01686, Sep. 2017.  
doi: \href{https://doi.org/10.48550/arXiv.1709.01686}{10.48550/arXiv.1709.01686}

% \bibitem{Hardware-Aware-Automated-Quantization} K. Wang, Z. Liu, Y. Lin, J. Lin, \& S. Han, \emph{HAQ: Hardware aware automated quantization with mixed precision}, in \emph{Proceedings of the IEEE/CVF Conference on Computer Vision and Pattern Recognition (CVPR)}, Long Beach, CA, USA, Jun. 2019, pp. 8604--8612. {https://doi.org/10.1109/CVPR.2019.00881}
\bibitem{Hardware-Aware-Automated-Quantization}
K. Wang, Z. Liu, Y. Lin, J. Lin, and S. Han,  
"HAQ: Hardware aware automated quantization with mixed precision,"  
in \emph{Proc. IEEE/CVF Conf. Comput. Vis. Pattern Recognit. (CVPR)}, Long Beach, CA, USA, Jun. 2019, pp. 8604--8612.  
doi: \href{https://doi.org/10.1109/CVPR.2019.00881}{10.1109/CVPR.2019.00881}

% \bibitem{Granularity-level} H. Mao, S. Han, J. Pool, and W. J. Dally, "Exploring the granularity of sparsity in convolutional neural networks," in \textit{Proc. IEEE Conf. Comput. Vis. Pattern Recognit. Workshops (CVPRW)}, Honolulu, HI, USA, Jul. 2017, pp. 1927--1934. doi: {https://doi.org/10.1109/CVPRW.2017.241}{10.1109/CVPRW.2017.241}.
\bibitem{Granularity-level}
H. Mao, S. Han, J. Pool, and W. J. Dally,  
"Exploring the granularity of sparsity in convolutional neural networks,"  
in \emph{Proc. IEEE Conf. Comput. Vis. Pattern Recognit. Workshops (CVPRW)}, Honolulu, HI, USA, Jul. 2017, pp. 1927--1934.  
doi: \href{https://doi.org/10.1109/CVPRW.2017.241}{10.1109/CVPRW.2017.241}

% \bibitem{credit-card-fraud-compare} I. Benchaji, S. Douzi, B. El Ouahidi, \& J. Jaafari, \emph{Enhanced credit card fraud detection based on attention mechanism and LSTM deep model}, \emph{Journal of Big Data}, vol. 8, no. 1, p. 151, Dec. 2021. {https://doi.org/10.1186/s40537-021-00541-8}
\bibitem{credit-card-fraud-compare}
I. Benchaji, S. Douzi, B. El Ouahidi, and J. Jaafari,  
"Enhanced credit card fraud detection based on attention mechanism and LSTM deep model,"  
\emph{J. Big Data}, vol. 8, no. 1, Art. no. 151, Dec. 2021.  
doi: \href{https://doi.org/10.1186/s40537-021-00541-8}{10.1186/s40537-021-00541-8}

% \bibitem{Some-studies-1} L. Yang, X. Shen, C. Zhong, \& Y. Liao, \emph{On-demand inference acceleration for directed acyclic graph neural networks over edge-cloud collaboration}, \emph{Journal of Parallel and Distributed Computing}, vol. 171, pp. 79--87, Jan. 2023. {https://doi.org/10.1016/j.jpdc.2022.09.005}
\bibitem{Some-studies-1}
L. Yang, X. Shen, C. Zhong, and Y. Liao,  
"On-demand inference acceleration for directed acyclic graph neural networks over edge-cloud collaboration,"  
\emph{J. Parallel Distrib. Comput.}, vol. 171, pp. 79--87, Jan. 2023.  
doi: \href{https://doi.org/10.1016/j.jpdc.2022.09.005}{10.1016/j.jpdc.2022.09.005}

% \bibitem{Some-studies-2} J. Tian, X. Li, \& X. Qin, \emph{Reinforcement learning based collaborative inference and task offloading optimization for cloud-edge-end systems}, in \emph{Proceedings of the 2024 International Joint Conference on Neural Networks (IJCNN)}, Yokohama, Japan, Jun. 2024, pp. 1--8. {https://doi.org/10.1109/IJCNN60899.2024.10651115}
\bibitem{Some-studies-2}
J. Tian, X. Li, and X. Qin,  
"Reinforcement learning based collaborative inference and task offloading optimization for cloud-edge-end systems,"  
in \emph{Proc. 2024 Int. Joint Conf. Neural Netw. (IJCNN)}, Yokohama, Japan, Jun. 2024, pp. 1--8.  
doi: \href{https://doi.org/10.1109/IJCNN60899.2024.10651115}{10.1109/IJCNN60899.2024.10651115}


\end{thebibliography}
\end{document}